%% file: main.tex
\documentclass[10pt,twocolumn,letterpaper]{article}
\PassOptionsToPackage{table}{xcolor}
\usepackage[pagenumbers]{iccv} % To force page numbers, e.g. for an arXiv version


\definecolor{iccvblue}{rgb}{0.21,0.49,0.74}
\usepackage[pagebackref,breaklinks,colorlinks,allcolors=iccvblue]{hyperref}

\usepackage{url}
\usepackage{booktabs}       % professional-quality tables
\usepackage{amsfonts}       % blackboard math symbols
\usepackage{nicefrac}       % compact symbols for 1/2, etc.
\usepackage{microtype}      % microtypography
\usepackage{enumitem} % used for itemize margins
\usepackage{graphicx}
\usepackage{multirow}
\usepackage{float}
\usepackage{multicol} 
\usepackage{amsmath}
\usepackage{array}
\usepackage{adjustbox}
\usepackage{soul}
\usepackage{wrapfig}
\usepackage{xspace}
\usepackage{comment}
\usepackage{colortbl}

\definecolor{iucolor}{HTML}{B0504D}
\definecolor{gcolor}{HTML}{5476A6}

\newcommand{\ourDataset}{SCOUT\xspace}

\definecolor{palegreen}{rgb}{0.85, 1.0, 0.85}   % Define palegreen color
\definecolor{palegray}{rgb}{0.95, 0.95, 0.95}
\definecolor{palered}{rgb}{1.0, 0.85, 0.85}
\definecolor{paleorange}{rgb}{1.0, 0.9, 0.7}

\definecolor{palepink}{rgb}{0.93, 0.9, 0.85}
\definecolor{palepeach}{rgb}{0.88, 0.94, 0.85}
\definecolor{palelime}{rgb}{0.85, 0.97, 0.85}

\definecolor{low}{RGB}{255, 102, 102}    % Red (0-30%)   % Lighter Red (0-30%)
\definecolor{medium}{RGB}{255, 204, 153} % Lighter Orange (30-60%)
\definecolor{high}{RGB}{153, 230, 153}   % Lighter Green (60-100%)

\def\paperID{10129} % *** Enter the Paper ID here
\def\confName{ICCV}
\def\confYear{2025}

\title{TWIST \& SCOUT: Grounding Multimodal LLM-Experts by Forget-Free Tuning}

\author{\textbf{Aritra Bhowmik}$^{*1}$ \hspace{5pt} \textbf{Mohammad Mahdi Derakhshani}$^{*1}$ \hspace{5pt} \textbf{Dennis Koelma}$^{1}$ \\
\textbf{Martin R. Oswald}$^{1}$ \hspace{5pt} \textbf{Yuki M. Asano}$^{2}$ \hspace{5pt} \textbf{Cees G. M. Snoek}$^{1}$ \\
$^{1}$University of Amsterdam \\
$^{2}$University of Technology Nuremberg \\
\thanks{Joint first authors. Corresponding authors: \{a.bhowmik, m.m.derakhshani\}@uva.nl}
}

\begin{document}
\maketitle
\input{sec/abstract}
\input{sec/introduction}

\input{sec/related_work}

\input{sec/methodology}

\input{sec/experiments}
\input{sec/conclusion}

{
    \small
    \bibliographystyle{ieeenat_fullname}
    \bibliography{main}
}

\newpage
\input{sec/appendix}

\end{document}

%% file: sec/abstract.tex
\begin{abstract}
Spatial awareness is key to enable embodied multimodal AI systems. Yet, without vast amounts of spatial supervision, current Multimodal Large Language Models (MLLMs) struggle at this task. In this paper, we introduce TWIST \& SCOUT, a framework that equips pre-trained MLLMs with visual grounding ability without forgetting their existing image and language understanding skills.
To this end, we propose TWIST, a twin-expert stepwise tuning module that modifies the decoder of the language model using one frozen module pre-trained on image understanding tasks and another learnable one for visual grounding tasks. This allows the MLLM to retain previously learned knowledge and skills, while acquiring what is missing.
To fine-tune the model effectively, we generate a high-quality synthetic dataset we call \ourDataset, which mimics human reasoning in visual grounding. This dataset provides rich supervision signals, describing a step-by-step multimodal reasoning process, thereby simplifying the task of visual grounding. We evaluate our approach on several standard benchmark datasets, encompassing grounded image captioning, zero-shot localization, and visual grounding tasks. Our method consistently delivers strong performance across all tasks, while retaining the pre-trained image understanding capabilities.
\end{abstract}

%% file: sec/introduction.tex
\section{Introduction}
\label{sec:intro}

Multimodal Large Language Models (MLLMs) have greatly advanced vision and language tasks, excelling in image captioning and visual question answering~\citep{alayrac2022flamingo, li2023blip, dai2024instructblip, liu2024visual}. Models like Flamingo, BLIP-2, InstructBLIP, and VisualGLM leverage large image-caption datasets to integrate vision and language, addressing complex multimodal challenges. However, due to their caption-based design, these models often lack visual grounding, limiting their suitability for tasks requiring precise spatial understanding~\citep{wen2023road, luo2024delving, driess2023palm, jin2023alphablock, cheng2024empowering}. While extensive pre-training can equip models with localization capabilities~\citep{wang2023cogvlm, chen2023shikra}, it requires massive datasets, human-annotated bounding boxes, and substantial computational resources, making it impractical for many setups. Instead, we focus on fine-tuning pre-trained MLLMs to instill spatial understanding in a forget-free manner, preserving existing language and vision comprehension skills.

\begin{figure*}[t]
    \centering
    \includegraphics[width=1\textwidth]{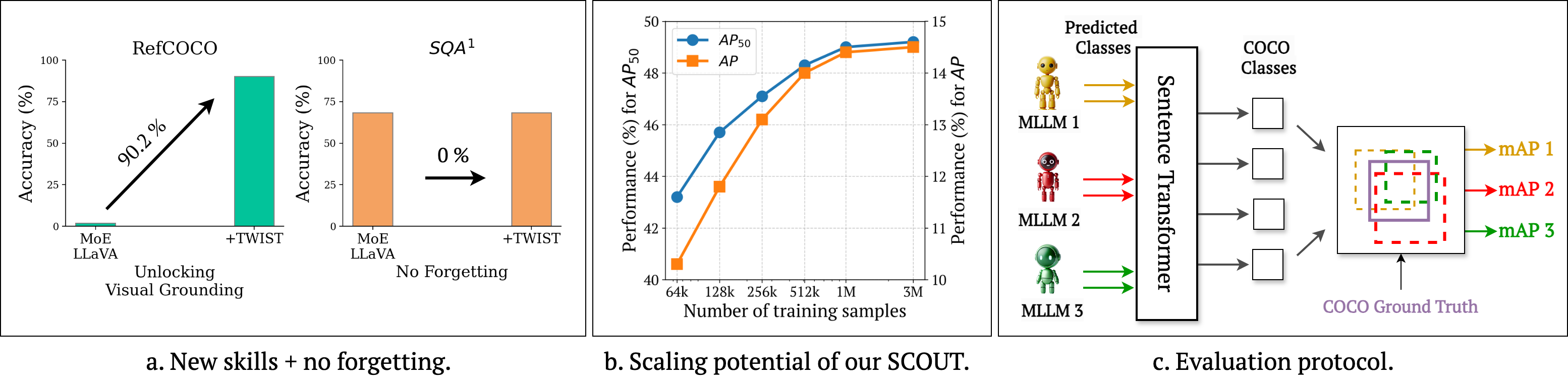}
    % \vspace{-12pt}
    \caption{\textbf{TWIST \& SCOUT contributions.} Our contributions include \textbf{(a)} TWIST, a framework that fine-tunes a pre-trained caption-based MLLM to acquire new grounding skills while retaining existing image understanding capabilities, \textbf{(b)} SCOUT, a scalable synthetic dataset that enhances model performance through step-by-step grounded chain-of-thought annotations, and \textbf{(c)} an evaluation protocol tailored for assessing MLLMs on grounded image captioning tasks.}
    
    % Our contributions include \textbf{(a)} enabling a pre-trained caption-based vision-language model to learn new grounding skills by fine-tuning without forgetting old skills, \textbf{(b)} improving model performance by simply scaling the generated dataset, and \textbf{(c)} enabling visual grounding tasks through step-by-step training on our synthetic dataset \cs{Shouldn't we include an image together with the text?}.}
    \vspace{-7pt}
    \label{fig:teaser}
\end{figure*}

% Closest to our work is PIN by Dorkenwald \etal~\citep{dorkenwald2024pin}, which addresses single-object localization in pre-trained autoregressive MLLMs through two key innovations: modifying the vision encoder with learned spatial parameters to enable bounding box prediction and introducing a synthetic dataset of superimposed object renderings to bypass the need for human-annotated data. However, PIN’s architectural modifications lead to catastrophic forgetting, erasing the model’s pre-trained image understanding capabilities. Additionally, the simplistic object-pasting approach of the dataset introduces domain shift and limits the model’s applicability to more complex tasks involving multi-object reasoning and richer spatial relationships~\citep{wang2023cogvlm, chen2023shikra}. Another potential approach is parameter-efficient tuning via LoRA~\citep{hu2021lora}, which adds low-rank weight updates to a frozen backbone. Although LoRA often preserves a model’s pre-trained strengths on tasks close to its original domain, its low-rank constraints and limited capacity struggle to capture new spatial relationships and bounding-box nuances, leading to suboptimal grounding performance. Consequently, neither PIN nor LoRA retains existing vision-language skills while adding robust grounding—an issue our work tackles without full model finetuning.

Closest to our work is PIN by Dorkenwald~\etal~\citep{dorkenwald2024pin}, which addresses single-object localization in pre-trained autoregressive MLLMs through two key innovations: modifying the vision encoder with learned spatial parameters for bounding box prediction and introducing a synthetic dataset of superimposed object renderings to remove reliance on human annotations. However, PIN’s architectural modifications cause catastrophic forgetting, erasing pre-trained image understanding. Additionally, its simplistic object-pasting approach introduces domain shift, limiting applicability to complex tasks requiring multi-object reasoning and richer spatial relationships~\citep{wang2023cogvlm, chen2023shikra}. Another approach is parameter-efficient tuning via LoRA~\citep{hu2021lora}, which adds low-rank weight updates to a frozen backbone. While LoRA preserves pre-trained strengths for tasks close to its domain, its low-rank constraints and limited capacity fail to capture new spatial relationships and bounding-box nuances, leading to suboptimal grounding. Consequently, neither PIN nor LoRA retains vision-language skills while adding robust grounding—an issue our work addresses without full model finetuning.

To equip autoregressive MLLMs with robust grounding while ensuring forget-free performance, we introduce TWIST $\&$ SCOUT. TWIST stands for \textbf{TWI}n-expert \textbf{S}tepwise \textbf{T}uning, a framework with two parallel modules and a stepwise loss function inspired by~\citet{lightman2023let}. We treat the pre-trained backbone as one “expert” and add a Mixture of Experts (MoE) as the second expert for grounding, providing enough capacity to handle unfamiliar demands without overwriting pre-trained understanding. Akin to LoRA, we add new parameters; however, rather than relying on low-rank residuals, we fuse old and new knowledge via a learnable gating mechanism, enabling robust grounding without erasing existing skills. Stepwise tuning strengthens learning by breaking down complex tasks into simpler subtasks, enhancing vision-language performance. Complementing TWIST, we present SCOUT, short for \textbf{S}ynthetic \textbf{C}hain-of-Th\textbf{ou}gh\textbf{t} with Grounding, a high-quality synthetic dataset capturing meaningful spatial relationships, inter-object reasoning, and stepwise thought processes—providing a rich training signal for fine-tuning MLLMs. Recognizing the limitations of evaluation methods focused solely on object localization, we introduce a protocol for assessing MLLMs on free-form grounded image captioning, which requires both visual grounding and image understanding. Our contributions can be summarized as:

\begin{enumerate} 
    \item We propose TWIST, a TWIn-expert Stepwise Tuning framework that fine-tunes pre-trained MLLMs via two parallel modules without forgetting. TWIST employs step-by-step training, breaking complex grounding tasks into simpler subtasks (Figure~\ref{fig:teaser} (a)).
    \item We present SCOUT, a synthetic dataset with stepwise grounded chain-of-thought annotations. SCOUT facilitates fine-tuning for grounding and reasoning, providing a rich, spatially complex training signal (Figure~\ref{fig:teaser} (b)).
    \item We create an evaluation protocol for assessing MLLMs on free-form grounded image captioning (Figure~\ref{fig:teaser} (c)).
\end{enumerate}
Our experiments show strong performance in grounded image captioning and visual grounding while retaining initial image understanding.

%% file: sec/related_work.tex
\section{Related Work}
\label{sec:related_work}

\noindent\textbf{Multimodal LLMs.}  
Large Language Models (LLMs), known for their instruction-following and generalization abilities, have been effectively integrated with vision encoders, achieving strong multimodal performance~\citep{alayrac2022flamingo,li2023blip, dai2024instructblip,bai2023qwen,lin2023video, wang2024visionllm, xue2024xgen, tong2024cambrian, mckinzie2024mm1, zhang2024mm1, deitke2024molmo, agrawal2024pixtral, li2024llava}. Pioneering models like Flamingo~\citep{alayrac2022flamingo} and BLIP-2~\citep{li2023blip} integrate vision and language by combining CLIP-based image encoders with LLMs—Flamingo using perceiver and gated cross-attention blocks, while BLIP-2 employs a lightweight Querying Transformer. Recent efforts have optimized training strategies~\citep{bai2023qwen, xue2024xgen}, improved image resolution~\citep{wang2023cogvlm, bai2023qwen, li2024llava}, and enhanced image encoders~\citep{chen2023internvl, zhang2023internlm}. Additional advancements refine input alignment~\citep{lin2023video} and projection layers~\citep{cha2023honeybee, dai2024instructblip}, while expanding instruction-tuning datasets has further improved performance and versatility~\citep{liu2023mitigating, zhang2023llavar}. However, despite these improvements, instruction-tuned MLLMs mainly excel at image captioning and simple QA but struggle with spatial reasoning and precise object grounding~\citep{dorkenwald2024pin}. Our work addresses these gaps by equipping MLLMs with spatial understanding for visual grounding and object localization.

\noindent\textbf{Grounded Multimodal Models and Object Detection.}  
Extending MLLMs beyond image and language understanding, several models have been developed to enable visual grounding and object localization~\citep{chen2021pix2seq, wang2022ofa, lu2022unified, yang2022unitab, wang2022git, wang2024visionllm, chen2023shikra, wang2023cogvlm, bai2023qwen}. Pix2Seq~\citep{chen2021pix2seq} pioneered treating object detection as an autoregressive language modeling task, inspiring models like OFA~\citep{wang2022ofa}, Unified-IO~\citep{lu2022unified}, UniTab~\citep{yang2022unitab}, and VisionLLM~\citep{wang2024visionllm} to integrate language and coordinate vocabularies for grounding. Meanwhile, Shikra~\citep{chen2023shikra}, CogVLM~\citep{wang2023cogvlm}, and Qwen-VL~\citep{bai2023qwen} further advance positional representations in natural language, facilitating seamless interleaved grounded captions. Despite these advancements, most models rely on large annotated datasets and extensive pre-training. Grounding DINO~\citep{liu2023grounding} takes a different approach, using a transformer-based architecture trained with contrastive and bounding box regression losses for object detection. However, unlike autoregressive MLLMs, Grounding DINO is optimized specifically for detection and lacks the ability to generate grounded image captions in free-form text. PIN~\citep{dorkenwald2024pin} attempts to bridge the gap by introducing a learnable positional insert module and a synthetic dataset for fine-tuning. Yet, its reliance on purely synthetic data leads to domain shift, causing it to forget previous vision-language abilities and remain limited to single-object localization. Our approach addresses these challenges through TWIST, a two-module framework that preserves pre-trained vision-language skills while incrementally adding grounding capabilities. Paired with SCOUT, our synthetic dataset featuring chain-of-thought reasoning, TWIST enables MLLMs to handle complex, multi-object grounding tasks requiring both spatial reasoning and image understanding.

%% file: sec/methodology.tex
% \begin{figure*}[t]
%     \centering
%     \includegraphics[width=0.8\textwidth]{figures/scout.png}
%     % \vspace{-12pt}
%     \caption{\textbf{SCouT pipeline}. We use LLMs to generate "where" and "what" questions from the input caption. With these questions and the image, we prompt large MLLMs to produce the SCouT grounding dataset, ensuring visually grounded and contextually relevant data.}
%     \vspace{-7pt}
%     \label{fig:scout}
% \end{figure*}

%\section{Synthetic Chain-of-Thought Grounding Data}

% To address the hallucination problem, we introduce the \textit{SCouT} dataset, which integrates visual information with textual descriptions from captions. We generate ``where'' and ``what'' questions using an LLM, such as Mixtral \cs{ref}, ensuring contextual alignment with captions through in-context prompting (see Appendix Figure \ref{fig:positve_prompts}). For answer generation, instead of relying solely on the LLM like Shikra, we use a pre-trained MLLM, such as CogMLLM, which incorporates image context to answer questions step-by-step, reducing hallucinations (see Appendix Figure \ref{fig:positive}). In a comparison of 100 randomly selected samples from each dataset, SCouT achieved an accuracy of 94.7\%, significantly outperforming Shikra’s 63.1\%. A correct response is defined as one where the generated grounding steps accurately reflect the relationships and spatial positions of objects in the image, without hallucinations. This demonstrates the effectiveness of our method in producing reliable, contextually grounded data.

\section{TWIST}
\label{sec:method}
In the following sections, we briefly review standard MLLMs and the concept of Mixture of Experts (MoE). We then introduce TWIST, a TWIn-expert Stepwise Tuning framework with two parallel modules and a step-by-step training objective. Finally, we explain how the step-by-step learning strategy adjusts the training loss.

\subsection{Preliminaries}

\noindent\textbf{Multimodal Large Language Models (MLLMs).}
MLLMs process both image and text data for multimodal generative tasks. These models consist of a vision encoder $\psi(\cdot)$, a language decoder, $\phi(\cdot)$, and a mapper function $f(\cdot)$. The language decoder takes a sequence of tokens as inputs $\left[v_{1}, v_{2}, \ldots, v_{m}, t_1, t_2, \ldots, t_{n}\right]$ being composed of visual and textual tokens. Visual tokens are computed from an image $\textbf{x}$ as $\left[v_1, v_2, \ldots, v_{m}\right] = f(\psi(\textbf{x}))$, and textual tokens are computed from the text input $\textbf{t}$ as $\left[t_1, t_2, \ldots, t_{n}\right] = \mathrm{Tokenizer}(\textbf{t})$. 
MLLMs are trained via the cross-entropy loss.

\noindent\textbf{Mixture of Experts (MoEs).}
MoEs are a way to increase small model capacity to compete with large models performance without a proportional increase in computational cost \citep{shazeer2017outrageously}. Specifically, an MoE layer is composed of $E$ ``experts" and a gating network $g(\cdot)$. The gating network decides which expert is most suitable for a given token:
\begin{equation}
    l_{n} = \texttt{MoE}(l_{n-1}) = \sum_{i=1}^{E} g_i(l_{n-1}) \cdot e_i(l_{n-1}),
    \label{eq:moe}
\end{equation}
where $l_{n}$ represents the output of the $n$-th layer, $l_{n-1}$ its input, $E$ the total number of experts, $g_i(\cdot)$ the gating function's weight for the $i$-th expert, and $e_i(\cdot)$ the $i$-th expert's output.
During inference, only the top-$k$ experts can be used, reducing inference costs considerably.

% In an MoE-based visual language models, a gating mechanism dynamically selects and weights the contribution of each expert based on the input data, effectively deciding which expert is most suitable for a given task. This selection process is formulated as:
% \begin{equation}
%     y = \sum_{i=1}^{E} g_i(x) \cdot e_i(x),
% \end{equation}
% where $y$ represents the output, $x$ the input, $E$ the total number of experts, $g_i(x)$ the gating function's weight for the $i$-th expert, and $e_i(x)$ the $i$-th expert's output.

% Training an MoE model involves optimizing both the experts and the gating mechanism, ensuring the model learns to effectively distribute tasks among its experts. This is done via cross-entropy loss like before, but here, the next token prediction is dependent on all the experts and the gating function.
% \begin{equation}
%     L_{MoE} = -\sum_{i=1}^{N} \log P_{\theta}(T_i | H_v, T_1, ..., T_{i-1}) + \lambda \cdot R(g),
% \end{equation}
% where $N$ represents the length of the text sequence, $T_i$ denotes the $i^{th}$ word in the sequence and now is given by $T_i = \sum_{i=1}^{E} g_i(T_{i-1}) \cdot e_i(T_{i-1})$, and $\theta$ symbolizes the model parameters. $\lambda$ a regularization coefficient, and $R(g)$ a regularization term for the gating mechanism.

\subsection{TWIST Workflow}
\begin{figure*}[t!]
    \centering
    \includegraphics[width=0.9\textwidth]{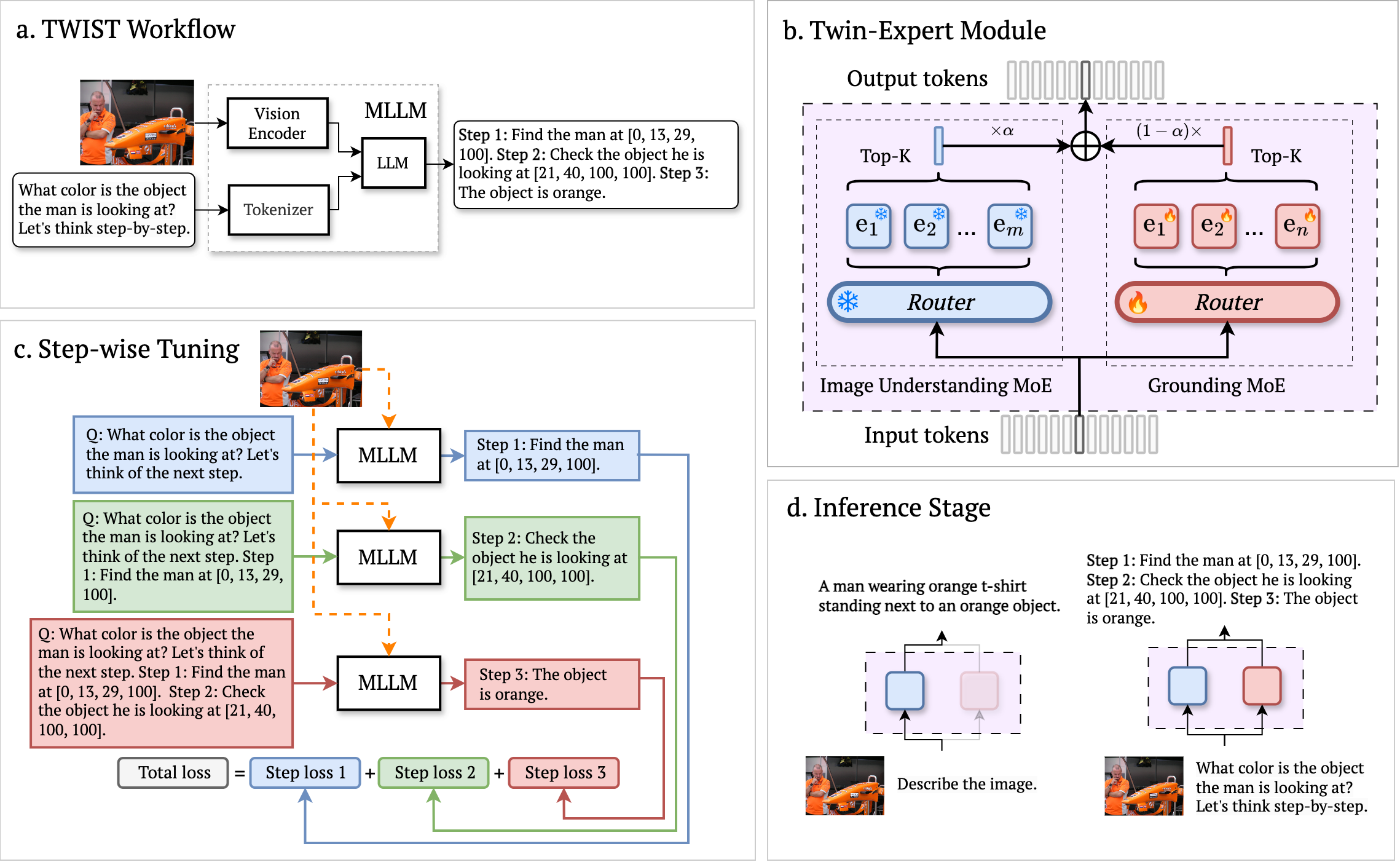}\\[-5pt]
    \caption{\textbf{TWIST system overview.} \textbf{(a)} The MLLM  processes an image and text prompt via a vision encoder and language decoder to generate outputs, \textbf{(b)} Twin-Expert, featuring two parallel mixture of experts modules: a frozen one for image understanding and a trainable one for visual grounding, \textbf{(c)} The stepwise loss function breaks down complex reasoning into sequential subtasks, simplifying the training process, \textbf{(d)} During inference, information flows through the image understanding module (blue box) for those tasks and through both modules (blue and red box) for grounding tasks.}
    \label{fig:arch}
\end{figure*}

In Figure \ref{fig:arch} (a), we present the general workflow of the TWIST model, which consists of a vision encoder, tokenizer and an LLM, taking image-text pairs as inputs and generating grounded free-form texts. Below, we detail each component of the TWIST workflow.

\noindent\textbf{Twin-expert module.}
We start with a caption-based mixture-of-expert MLLM~\citep{lin2024moe} adept at visual question answering tasks, and extend it for the task of visual grounding as depicted in Figure \ref{fig:arch} (b). 
A transformer block of the language decoder of a MLLM is composed of multi-head attention (MHA), a feed-forward network (FFN) and a layer norm (LN), which processes the input tokens as follows:
\begin{align}
    \hat{l}_{n} & = \texttt{MHA}(\texttt{LN}(l_{n-1})) + l_{n-1}, \label{eq:attention_block}\\ 
% \end{equation}
% \begin{equation}
    l_{n} &= \texttt{FFN}(\texttt{LN}(\hat{l}_{n})) + \hat{l}_{n},    \label{eq:ffn_block}
\end{align}
where $l_{n-1}$ is the input from layer $n{-}1$, $\hat{l}_{n}$ is the hidden representation at layer $n$, and $l_{n}$ is the output of the $n$-th layer.
The mixture of expert module only modifies Eq.~\eqref{eq:ffn_block} by replacing the FFN module with a MoE in the transformer block computation as follows:
\begin{equation}
    l_{n} = \texttt{MoE}(\texttt{LN}(\hat{l}_{n})) + \hat{l}_{n}.
    \label{eq:moe_block}
\end{equation}
We introduce a parallel MoE module for visual grounding and modify the above equations as follows:
% \begin{equation}
%     l_{n}^{\texttt{\color{iucolor}IU}} = \texttt{MoE}^{\texttt{{\color{iucolor}IU}}}(\texttt{LN}(\hat{l}_{n})) + \hat{l}_{n},  \; \; \; \; \;  l_{n}^{\texttt{\color{gcolor}VG}} = \texttt{MoE}^{\texttt{\color{gcolor}VG}}(\texttt{LN}(\hat{l}_{n})) + \hat{l}_{n}, 
%     \label{eq:lynx_block_iu}
% \end{equation}
\begin{equation}
\begin{split}
    l_{n}^{\texttt{\color{iucolor}IU}} &= \texttt{MoE}^{\texttt{{\color{iucolor}IU}}}(\texttt{LN}(\hat{l}_{n})) + \hat{l}_{n}, \\
    l_{n}^{\texttt{\color{gcolor}VG}} &= \texttt{MoE}^{\texttt{\color{gcolor}VG}}(\texttt{LN}(\hat{l}_{n})) + \hat{l}_{n},
\end{split}
\label{eq:lynx_block_iu}
\end{equation}
% \begin{equation}
%     l_{n+1}^{\texttt{\color{gcolor}G}} = \texttt{MoE}_{\texttt{\color{gcolor}G}}(\texttt{LN}(l_{n})) + l_{n}, 
%     \label{eq:lynx_block_grounding}
% \end{equation}
\begin{equation}
    l_{n} = \alpha \cdot l_{n}^{\texttt{\color{iucolor}IU}} + (1-\alpha) \cdot l_{n}^{\texttt{\color{gcolor}VG}},
    \label{eq:lynx_block}
\end{equation}
where $\texttt{MoE}^{\texttt{\color{iucolor}IU}}$ is a frozen MoE module pre-trained on image understanding tasks, $\texttt{MoE}^{\texttt{\color{gcolor}VG}}$ is a learnable MoE module trained for visual grounding task, and $\alpha$ is a learnable coefficient weight adjusting the contribution of each MoE module. 
This design choice prevents catastrophic forgetting of pre-trained image understanding skills of MLLMs. Moreover, the shared modules allow knowledge transfer from the pre-trained image understanding MoE into the grounding MoE, helping the latter to better interpret grounding tasks.

\noindent\textbf{Training step.} We train our model using a cross-entropy loss for the next token prediction task:
\begin{equation}
    L = -\left[\sum_{i=1}^{N} \log P_{\theta}(t_i | v_{1}, \ldots, v_{m}, t_1, \ldots, t_{i-1})\right] + \lambda \cdot R(g),
    \label{eq:loss_function}
\end{equation}
%
%\mo{the scope and shape variable g is not really clear here}
%
where $L$ is the next token prediction loss, $N$ represents the length of the text sequence, $v_i$ refers to the $i$-th visual token in the sequence, $t_i$ denotes the $i$-th textual token in the sequence, $\theta$ refers to the model parameters, $\lambda$ is a regularization coefficient, and $R(g)$ is a regularization term for sparsifying the gating mechanism. This loss function aims to minimize the discrepancy between the predicted and actual next token in the sequence.

\noindent\textbf{Step-by-step loss function.} To fully leverage the Twin-Expert module of TWIST, we implement a step-by-step loss inspired by \citet{lightman2023let}. This approach decomposes complex tasks into sequential, easily digestible subtasks, each corresponding to a specific part of the overall reasoning process, as seen in Figure~\ref{fig:arch} (c). These steps are not separate tasks but subtasks of a unified task. To illustrate this concept mathematically, the loss function for training under step-by-step reasoning supervision can be expressed as:

% \begin{equation} 
%     L_{\text{step-by-step}} = \sum_{j=1}^{J} \left[ - \left( \sum_{i=1}^{N_j} \log P_{\theta}(t_{i}^{(j)} \mid v_{1}, \ldots, v_{m}, t_{1}^{(j)}, \ldots, t_{i-1}^{(j)}) \right) \right] + \lambda \cdot R(g) ,
% \end{equation}. 
\begin{equation}
\begin{split}
    \mathcal{L}_{\text{step-by-step}} = \sum_{j=1}^{J} \Bigg[ - \Bigg( \sum_{i=1}^{N_j} \log P_{\theta}(t_{i}^{(j)} \mid v_{1}, \ldots, \\ v_{m}, t_{1}^{(j)}, \ldots, t_{i-1}^{(j)}) \Bigg) \Bigg]
    + \lambda \cdot R(g) ,
\end{split}
\end{equation}
where $L_{\text{step-by-step}}$ represents the step-by-step reasoning loss function, $J$ is the number of reasoning steps, ${N_j}$ is the number of tokens in step $j$, $t_{i}^{(j)}$ represents the $i^{th}$ token in the $j^{th}$ step output, $v_{1}...v_{m}$ are the image tokens, $P_{\theta}$ is the probability predicted by the model and $R(g)$ is the regularization term with weight $\lambda$.

\noindent\textbf{Inference step.} During inference, we determine the task type (image understanding or visual grounding) based on the input prompt and adjust $\alpha$ accordingly. We employ a lightweight BERT-based classifier \citep{devlin2018bert} which takes an input prompt and classifies it into one of the two task categories. Based on the classifier's output, $\alpha$ is adjusted dynamically: 
\begin{equation}
\alpha = 
\begin{cases} 
1 & \text{for Image Understanding}, \\
\text{unchanged} & \text{for Visual Grounding}.
\end{cases}
\end{equation}
Thus, at test time, the output of the twin-expert module, as depicted in Figure~\ref{fig:arch} (d), is as follows: 
\begin{equation}
l_{n+1} \!=\! 
\begin{cases} 
l_{n+1}^{\texttt{\color{iucolor}IU}} & \!\text{for Image Understanding}, \\
\alpha \!\cdot\! l_{n+1}^{\texttt{\color{iucolor}IU}} \!+ (1\!-\!\alpha) \!\cdot\! l_{n+1}^{\texttt{\color{gcolor}VG}} & \!\text{for Visual Grounding}.
\end{cases}
\end{equation}

The BERT classifier adds minimal computational overhead, as it is an 8-bit quantized tiny model with approximately 1 million parameters, bringing the total active parameters from 1.67B to 1.671B. Our experiments show that the classifier achieves 99.98\% accuracy, ensuring negligible impact on performance.

\section{SCOUT}
\label{ScouT}

% \noindent\textbf{Preliminary.}
% Visual question answering datasets often incorporate spatial reasoning, such as \textit{``What object is to the left of the girl?"} or \textit{``Is there a bowl on top of the table?"}. Grounding tasks particularly benefit from this spatial reasoning, as describing relationships like ``A cat at [x1, y1, x2, y2] sits to the left of a dog at [a1, b1, a2, b2]" offers clearer relative positioning, improving the interpretation of localization data for MLLMs. Following this intuition, recent works like Shikra~\citep{chen2023shikra} have made progress in creating grounded chain-of-thought multimodal datasets. Shikra uses an LLM to generate reasoning-based question-answer pairs from image captions, without access to the actual visual content. 
% However, this reliance on captions alone leads to hallucinated narratives that do not reflect the image (see hallucination examples of the Shikra dataset in Figure A.3 of our Appendix).

\noindent\textbf{Preliminaries.}  
Visual question answering datasets often involve spatial reasoning, such as \textit{``What object is to the left of the girl?"} or \textit{``Is there a bowl on top of the table?"}. Grounding tasks benefit from this reasoning, as describing relationships like ``A cat at [x1, y1, x2, y2] sits to the left of a dog at [a1, b1, a2, b2]" provides clearer relative positioning, improving localization interpretation for MLLMs. Recent works like Shikra~\citep{chen2023shikra} have explored grounded chain-of-thought multimodal datasets, using LLMs to generate reasoning-based Q\&A pairs from image captions—without direct visual access. However, relying solely on captions leads to hallucinated narratives that fail to reflect the actual image (see hallucination examples of the Shikra dataset in Figure A.3 of our Appendix).

\begin{figure}[t]
    \centering
    \includegraphics[width=1\linewidth]{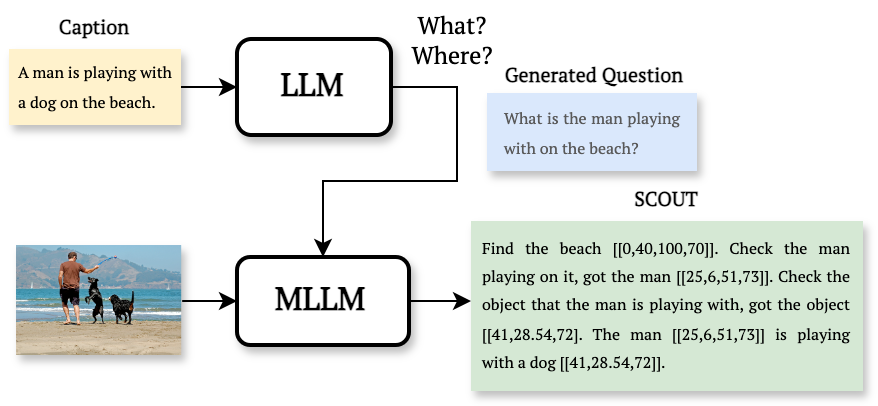}
    % \vspace{-12pt}
    \caption{\textbf{SCOUT data generation}. We use an LLM to generate ``what" and ``where" questions from the input caption. With these questions and the image, we prompt an MLLM to produce the SCOUT grounding dataset, ensuring visually grounded and contextually relevant data.}
    \vspace{-12pt}
    \label{fig:scout}
\end{figure}

\noindent\textbf{SCOUT data generation.} To generate our \ourDataset dataset and ensure high-quality, visually grounded data, we adopt a two-step process designed to mitigate the biases of text-only methods (see Figure \ref{fig:scout}). %
We begin by taking an image-caption pair from the Flickr30k \citep{plummer2015flickr30k} dataset, then use an LLM like Mixtral~\citep{jiang2024mixtral} with in-context prompting to generate ``what" and ``where" type spatial reasoning questions, focused on objects mentioned in the captions. This ensures the questions are relevant and grounded in the initial textual description. Additionally, we create negative samples by generating questions about objects that are not present in the image. These negative samples train the model to identify when queries are invalid or irrelevant. See Figures A.5 and A.6 in the Appendix.
To reduce hallucinations—specifically, the kind caused by relying solely on text descriptions without verifying or aligning with the actual visual content—we use a state-of-the-art MLLM, CogVLM~\citep{wang2023cogvlm}, for answer generation. CogVLM is recognized for its strong performance in visual grounding and reasoning tasks, making it a reliable choice for generating high-quality, contextually accurate answers. Since we rely on CogVLM for generating \ourDataset, the quality of our data—and consequently, our model's upper bound performance—is inherently tied to CogVLM's reasoning capabilities. For the positive samples, we feed CogVLM the image and the question, prompting it to analyze the visual scene step-by-step, ensuring the answers accurately reflect objects and spatial relationships present in the image. For the negative samples, we skip feeding them to the MLLM and instead explicitly indicate that the referenced object is not in the image. This approach helps the model learn to distinguish valid questions from those that are irrelevant or incorrect, thereby enhancing overall accuracy and robustness.

\noindent\textbf{SCOUT data quality.} In a small-scale human analysis of 100 randomly selected samples, \ourDataset achieved an accuracy of 94.7\%, significantly outperforming Shikra’s 63.1\%. A response was considered correct if the predicted object relationships and spatial positions matched the ground truth with at least 50\% Intersection over Union (IoU) for bounding boxes and accurately described the relative spatial arrangement between objects based on the image content.

%% file: sec/experiments.tex
\begin{table*}[t]
\centering
% \vspace{0.2cm}
\resizebox{\textwidth}{!}{%
\begin{tabular}{ll ccc | ccc | ccc }
\toprule
& & \multicolumn{3}{c}{$\textbf{PVOC}_{\leq 3 \text{ Objects}}$ } & \multicolumn{3}{c}{$\textbf{COCO}_{\leq 3 \text{ Objects}}$ } & \multicolumn{3}{c}{$\textbf{LVIS}_{\leq 3 \text{ Objects}}$ }\\
\cmidrule(r){3-5} \cmidrule(r){6-8} \cmidrule(r){9-11}
Method & Model & mIoU & $\text{mIoU}_M$ & $\text{mIoU}_L$ & mIoU & $\text{mIoU}_M$ &  $\text{mIoU}_L$ &  mIoU & $\text{mIoU}_M$ &  $\text{mIoU}_L$  \\
\midrule

PIN & OpenFlamingo & 0.45 & 0.27 & 0.62 & 0.35 & 0.26 &  0.59 &  0.26&  0.24&  0.61 \\
LoRA & OpenFlamingo & 0.44 & 0.26 & 0.62 & 0.33 & 0.23 &  0.58 &  0.23 & 0.19 & 0.55 \\
% PIN w/ BLIP-2~\cite{dorkenwald2024pin} &  0.44  & 0.24 &  0.63 & 0.34& 0.22 & 0.60 &  0.26 &  0.23 &  0.60\\
LoRA & MoE-LLaVA   &  0.43  & 0.21 &  0.65 & 0.36 & 0.29 & 0.60 &  0.24 &  0.21 &  0.62\\
\rowcolor{high}
TWIST \& SCOUT & MoE-LLaVA &  0.68  & 0.58 &  0.81 & 0.66 & 0.57 & 0.78 &  0.65 &  0.55 &  0.76 \\
\bottomrule

\end{tabular}
}
\caption{\textbf{Object localization comparison} with PIN~\citep{dorkenwald2024pin} and LoRA~\citep{hu2021lora} on three benchmarks. TWIST consistently outperforms PIN across various datasets and metrics. Although PIN and TWIST use different backbones, making direct comparisons tricky, the LoRA variants peform on par, but TWIST shows a much larger improvement over its LoRA variant compared to PIN.}
\label{tab:pin_comparison}
\vspace{-5mm}
\end{table*}

\section{Experiments}
\label{sec:experiments}
We evaluate our approach on three grounding tasks: 
i) object localization, ii) grounded image captioning, and iii) visual grounding, as well as standard image understanding tasks. Below, we detail our architectural implementation and training datasets.

\noindent{\textbf{Implementation details.}}  
Our twin-expert module is built on MoE-LLaVA~\citep{lin2024moe}, which uses Phi-2 as its pre-trained language model. MoE-LLaVA has four experts for image understanding tasks, and we add a separate MoE with two experts for grounding, initialized from the image understanding MoE in the same decoder layer. The vision encoder and multi-head attention layers remain frozen, as do the interleaved decoder layers. We optimize the model with AdamW~\citep{loshchilov2018decoupled} using a $2e{-}5$ learning rate, training on four \text{A}6000 GPUs for 1.5 days. The model contains 1.67B trainable parameters, with 0.8B active. We will release the code and our synthetic datasets.

\noindent{\textbf{Datasets.}}  
In addition to \ourDataset, we train our model on two established datasets. We use the RefCOCO dataset~\citep{yu2016modeling}, comprising three splits—RefCOCO, RefCOCO+, and RefCOCO$_\text{g}$—with a total of $128{,}000$ image-referential expression pairs from COCO$2014$~\citep{lin2014microsoft}. We also use $108k$ images from COCO$2017$, excluding the $6549$ unique RefCOCO val/test images to prevent data leakage. Since COCO provides only object labels and bounding boxes, we use CogVLM to generate grounded image captions, forming the GIC dataset. TWIST \& SCOUT refers to TWIST trained on REC, GIC, and SCOUT, using $512k$ samples from \ourDataset unless otherwise specified.

\subsection{Object Localization}
\noindent\textbf{Setup.} We evaluate single-object localization—a core grounding capability—by comparing TWIST \& SCOUT with PIN~\citep{dorkenwald2024pin} and LoRA~\citep{hu2021lora} in Table~\ref{tab:pin_comparison}. PIN's evaluation requires generating bounding boxes when prompted with object names. Their evaluation is conducted on \textit{subsets} of COCO~\citep{lin2014microsoft}, Pascal VOC (PVOC)~\citep{everingham2010pascal}, and LVIS~\citep{gupta2019lvis}, with up to three objects per image, totaling 3,582, 2,062, and 6,016 test images, respectively. The mean Intersection over Union (mIoU) is reported for all bounding boxes, along with separate scores for medium ($32 \times 32$ to $96 \times 96$ pixels) and large (over $96 \times 96$ pixels) objects, quantifying overlap between predicted and true boxes.

\noindent\textbf{Results.} Although TWIST and PIN use different backbones—complicating direct comparisons—Table~\ref{tab:pin_comparison} shows that TWIST \& SCOUT outperforms PIN trained on the OpenFlamingo~\citep{awadalla2023openflamingo} backbone in single-object localization, improving mIoU by $22\%$ on PVOC, $32\%$ on COCO, and $39\%$ on LVIS, particularly excelling with medium objects. Meanwhile, fine-tuning MoE-LLaVA via LoRA underperforms across all datasets, reinforcing the need for our approach. Notably, TWIST’s improvement over its LoRA counterpart exceeds that of PIN over its own LoRA variant, demonstrating TWIST’s superior adaptability to new grounding tasks without erasing pre-trained vision-language expertise.

\begin{table*}[h]
\centering
\resizebox{0.87\textwidth}{!}{%
\begin{tabular}{lccccccccc}
\toprule
 & & & \multicolumn{3}{c}{\textbf{RefCOCO}} & \multicolumn{3}{c}{\textbf{GIC}} \\
\cmidrule(r){4-6} \cmidrule(r){7-9} 
 Method & Parameters & Type & val & test-A & test-B  & AP & AP$_{50}$ & AP$_{L}$ \\
\midrule
 Shikra-7B~\citep{chen2023shikra} & 7.0B & pre-trained & \cellcolor{medium} 87.0 & \cellcolor{medium} 90.6 & \cellcolor{medium} 80.2 & \cellcolor{medium} 13.2 & \cellcolor{medium} 46.8 & \cellcolor{medium} 16.7 \\
 Grounding DINO~\citep{liu2023grounding} & 172M & pre-trained & \cellcolor{high} 90.6 & \cellcolor{high} 93.2 & \cellcolor{high} 88.2 & \cellcolor{low} 0 & \cellcolor{low} 0 & \cellcolor{low} 0 \\
 Ferret-7B~\citep{you2023ferret} & 7.0B & pre-trained & \cellcolor{medium} 87.5 & \cellcolor{medium} 91.3 & \cellcolor{medium} 82.4 & \cellcolor{medium} 13.9 & \cellcolor{medium} 47.1 & \cellcolor{medium} 17.4 \\
 \midrule
 PIN~\citep{dorkenwald2024pin} & 1.2M & fine-tuned & \cellcolor{low} n.a. & \cellcolor{low} 26.4 & \cellcolor{low} n.a. & \cellcolor{low} 0 & \cellcolor{low} 0 & \cellcolor{low} 0 \\
 TWIST \& SCOUT & 1.6B & fine-tuned & \cellcolor{medium} 87.2 & \cellcolor{medium} 90.2 & \cellcolor{medium} 80.3 & \cellcolor{high} 15.0 & \cellcolor{high} 49.3 & \cellcolor{high} 19.1 \\
 \bottomrule
\end{tabular}}
\vskip -0.05in
\caption{\textbf{Visual grounding.} Object detectors like Grounding DINO excel in REC but fail in GIC, while pre-trained models like Ferret-7B and Shikra-7B perform moderately in both. TWIST \& SCOUT bridges this gap, achieving the best GIC performance while maintaining strong REC results, demonstrating the benefit of incremental fine-tuning over full retraining. Note that \textbf{\textcolor{low}{red}} indicates failure, \textbf{\textcolor{medium}{orange}} represents moderate performance, and \textbf{\textcolor{high}{green}} highlights the best performance.}

\label{tab:merged_visual_grounding}
\vskip -0.1in
\end{table*}

\begin{table*}[h]
\centering
  \resizebox{\textwidth}{!}{%
\begin{tabular}{lrccccccc}
\toprule
& & \multicolumn{3}{c}{\textbf{Image Question Answering}} & \multicolumn{4}{c}{\textbf{Benchmark Toolkit}} \\
\cmidrule(r){3-5} \cmidrule(r){6-9}
Method & Parameters & $\text{GQA}$ & $\text{SQA}^{\text{1}}$ & $\text{VQA}^{\text{T}}$ & $\text{POPE}$ & $\text{MME}$ & $\text{LLaVA}^{\text{W}}$ & $\text{MM-Vet}$ \\

\midrule
PIN \citep{dorkenwald2024pin}     & 1.2M & n.a. & n.a. & n.a. & n.a. & n.a. & n.a. & n.a. \\
% I-80B \citep{laurenccon2024obelics}     & 65.0B & 45.2 & - & 30.9 & - & - & - & - \\
LLaVA-phi2 & 13.0B & -- & 68.4 & 48.6 & 85.0 & 1335.1 & -- & 28.9 \\
MoE-LLaVA-phi2 (our base) & 3.6B & 61.4 & 68.5 & 51.4 & 86.3 & 1423.0 & 94.1 & 34.3 \\
\rowcolor{high}
TWIST \& SCOUT      & 1.6B & 61.4 & 68.5 & 51.4 & 86.3 & 1423.0 & 94.1 & 34.3 \\
 \bottomrule
\end{tabular}}
\vskip -0.05in
\caption{\textbf{Image understanding comparison.} 
We retain the image understanding abilities of our base model (MoE-LLaVA) through the twin-expert step-wise tuning framework, while PIN fails in image understanding tasks. Note that ``n.a.'' denotes that the corresponding method is inherently incapable of performing the specified task and ``--'' means the numbers are not reported by the baselines. 
}
\label{tab:image_understanding}
\vskip -0.2in
\end{table*}

% \subsection{Visual Grounding}
% \noindent\textbf{Setup.} We evaluate TWIST \& SCOUT on two key visual grounding tasks: referential expression comprehension (REC) and phrase grounding in Table \ref{tab:visual_grounding}. For REC, we evaluate on RefCOCO, RefCOCO+, and RefCOCOg~\citep{yu2016modeling} datasets, where the objective is to identify a single object in an image based on a descriptive query. In contrast, phrase grounding, evaluated on the Flickr30k Entities~\citep{plummer2015flickr30k} dataset, involves linking multiple objects to their corresponding noun phrases in a sentence, requiring more complex contextual reasoning.

% \noindent\textbf{Results.} Unlike most models in this comparison, which rely on extensive pre-training for grounding, both TWIST \& SCOUT and PIN are fine-tuning approaches. However, while PIN struggles with this task, TWIST \& SCOUT remains highly competitive with the best pre-trained models, achieving a 1.4\% improvement over Shikra-7B on the Flickr30k phrase grounding task. More comparisons in Appendix (Table \ref{tab:protocol_extra}).

\subsection{Visual Grounding}
\noindent\textbf{Setup.} We compare our models to existing literature on the following two types of visual grounding tasks:

\noindent\textbf{$\triangleright$ Grounded Image Captioning.}  
Grounded image captioning extends object detection by requiring models to recognize and localize objects within free-form text. Unlike standard detection tasks with predefined categories, this task generates structured outputs while aligning textual and visual elements. The lack of a standard evaluation protocol complicates model comparisons. To address this, we propose a protocol that maps object names from different MLLMs to COCO class labels using a sentence transformer~\citep{reimers2019sentence} (Figure~\ref{fig:teaser} (c)). We then evaluate models with COCO-style metrics, leveraging standardized annotations for consistency and fairness.

\noindent\textbf{$\triangleright$ Referential Expression Comprehension.}  
Referential expression comprehension (REC) focuses on identifying a single object in an image based on a descriptive query. We evaluate this task using the RefCOCO~\citep{yu2016modeling} dataset, where models must accurately localize the target object given natural language descriptions which requires a deeper understanding of contextual relationships.

\noindent\textbf{Results.}  
Table~\ref{tab:merged_visual_grounding} compares models on referential expression comprehension (REC) and grounded image captioning (GIC), highlighting their strengths and limitations. For REC, Grounding DINO \cite{liu2023grounding} achieves the highest accuracy (green), as expected for a specialized object detector, while Ferret-7B \cite{you2023ferret} and Shikra-7B \cite{chen2023shikra} perform competitively (orange) due to large-scale pre-training. TWIST \& SCOUT remains on par, showing that fine-tuning preserves strong grounding capabilities, whereas PIN underperforms (red), revealing the limitations of its synthetic training. For GIC, Grounding DINO fails entirely (red) due to its lack of language capabilities. TWIST \& SCOUT achieves the best performance (green), surpassing Ferret-7B by 2.2 $\text{AP}_{50}$, reinforcing the advantage of fine-tuning VLMs for multi-object grounding. While Ferret-7B and Shikra-7B perform well (orange), they still fall short, showing that pre-training alone is insufficient for mastering both spatial and semantic reasoning. Check Table A.3 and A.4 in Appendix for full comparison. 

These results confirm our core hypothesis: models trained for one task struggle with another. Grounding DINO excels in REC but fails in GIC, while Ferret-7B and Shikra-7B perform moderately in both but do not surpass our fine-tuned approach. TWIST \& SCOUT bridges this gap, adding grounding abilities to MLLMs while preserving vision-language understanding—without full retraining.

\subsection{Image Understanding}
\noindent\textbf{Setup.} An appealing characteristic of TWIST \& SCOUT is its ability to retain image understanding capabilities even after fine-tuning for grounding tasks. 

\noindent\textbf{Results.} As shown in Table~\ref{tab:image_understanding}, our approach matches the performance of MoE-LLaVA (our base) and is better than much larger models like LLaVA-phi2~\citep{liu2024visual}, despite being nearly ten times smaller. The reported numbers, except for MME, reflect accuracy scores, while MME represents a cumulative perception score with a maximum value of 2000.

\subsection{Ablations}

\noindent\textbf{Component ablation.} Table~\ref{tab:component_ablation} breaks down the contribution of each component in TWIST \& SCOUT. Without TWIST, the model completely lacks image understanding, as reflected in the $0$ MM-Vet score. Introducing TWIST restores image understanding ($34.3$ MM-Vet) while slightly improving grounding performance ($+0.8$ in RefCOCO, $+1.8$ in COCO), indicating that retaining pre-trained knowledge benefits grounding to some extent. Adding SCOUT further enhances grounding, boosting RefCOCO by $1.3$ and COCO by $2.3$, confirming its role in improving spatial reasoning. Finally, applying step-wise loss leads to the best performance, particularly on COCO ($+1.3$), showing that structured learning helps integrate SCOUT’s knowledge more effectively.
\begin{table}[h]
\centering
% \vskip -0.2in
\setlength{\tabcolsep}{2.1pt}
\renewcommand{\arraystretch}{0.65}
\begin{tabular}{cccccc}
\toprule
TWIST & SCOUT & Step-wise & MM-Vet & RefCOCO & COCO  \\
\midrule
\texttimes & \texttimes & \texttimes & 0 & 84.8 & 9.6\\
$\checkmark$ & \texttimes & \texttimes & 34.3 & 85.6 & 11.4\\
$\checkmark$ & $\checkmark$ & \texttimes & 34.3 & 86.9 & 13.7\\
% \texttimes & $\checkmark$ & $\checkmark$ & 0 & 56.4 & 8.1\\
\rowcolor{high}
$\checkmark$ & $\checkmark$ & $\checkmark$ & 34.3 & 87.2 & 15.0\\
\bottomrule
\end{tabular}
\caption{\textbf{Component ablation.} TWIST preserves image understanding (MM-Vet), SCOUT enhances grounding abilities, and the step-wise loss simplifies SCOUT, making grounding easier to learn.}
\label{tab:component_ablation}
\vskip -0.1in
\end{table}

\begin{table*}[t]
\centering
  \resizebox{0.95\textwidth}{!}{%
\begin{tabular}{lccccccccc}
\toprule
& \multicolumn{2}{c}{\textbf{Datasets}} & \multicolumn{3}{c}{\textbf{Question Answering}} & \multicolumn{3}{c}{\textbf{Visual Grounding}} \\
\cmidrule(r){2-3} \cmidrule(r){4-6} \cmidrule(r){7-9}
Method & LLaVA-mix-665k  & SCOUT & $\text{GQA}$ & $\text{SQA}$ & $\text{VQA}^{\text{T}}$ & $\text{AP}$ & $\text{AP}_{\text{50}}$ & $\text{AP}_{\text{L}}$ \\
\midrule % $\checkmark$ & \texttimes
\multirow{3}{*}{MoE-LLaVA} 
 & $\checkmark$ & \texttimes & 61.4 & 68.5 & 51.4 & 0    & 0    & 0   \\
 & $\checkmark$ & $\checkmark$  & 53.1 & 56.9 & 46.3 & 8.1  & 32.6 & 10.3 \\
 & \texttimes & $\checkmark$  & 0    & 0    & 0    & 10.7 & 35.2 & 12.9 \\
\midrule
\rowcolor{high}
TWIST & \texttimes & $\checkmark$     & 61.4 & 68.5 & 51.4 & 15.0 & 49.3 & 19.1 \\
 \bottomrule
\end{tabular}}
\vskip -0.05in
\caption{\textbf{Fine-tuning challenges for image understanding and grounding tasks.} Fine-tuning on one task leads to catastrophic forgetting of the other, while joint fine-tuning remains suboptimal. TWIST \& SCOUT preserves both abilities effectively.}
\label{tab:finetune_combined}
\vskip -0.2in
\end{table*}

\begin{table}[h]
\centering
\resizebox{0.74\linewidth}{!}{%
\begin{tabular}{lcc}
\toprule 
Methods         & $\text{VQA}^{\text{T}}$ & VQA-RAD \\ \midrule
MoE-LLaVA       & 31.7 & 28.5 \\
\rowcolor{high}
TWIST  & 51.4 & 63.1 \\
\bottomrule
\end{tabular}
}
\vskip -0.05in
\caption{\textbf{Fine-tuning with domain shift.} Adding the biomedical VQA-RAD degrades MoE-LLaVA’s performance, while TWIST \& SCOUT maintains strong results across all tasks.}
% \vskip -0.2in
\label{tab:biomedical_ft}
\end{table}

% \begin{table*}[t]
% \centering
%   \resizebox{0.95\textwidth}{!}{%
% \begin{tabular}{lccccccccc}
% \toprule
% & \multicolumn{2}{c}{\textbf{Datasets}} & \multicolumn{3}{c}{\textbf{Image Question Answering}} & \multicolumn{4}{c}{\textbf{Visual Grounding}} \\
% \cmidrule(r){2-3} \cmidrule(r){4-6} \cmidrule(r){7-9}
% Method & LLaVA-mix-665k  & SCOUT & $\text{GQA}$ & $\text{SQA}$ & $\text{VQA}^{\text{T}}$ & $\text{AP}$ & $\text{AP}_{\text{50}}$ & $\text{AP}_{\text{L}}$ \\
% \midrule % $\checkmark$ & \texttimes
% MoE-LLaVA & $\checkmark$ & \texttimes & 61.4 & 68.5 & 51.4 & 0    & 0    & 0   \\
%  & $\checkmark$ & $\checkmark$  & 53.1 & 56.9 & 46.3 & 8.1  & 32.6 & 10.3 \\
%  & \texttimes & $\checkmark$  & 0    & 0    & 0    & 10.7 & 35.2 & 12.9 \\
% \midrule
% \rowcolor{high}
% TWIST \& SCOUT & \texttimes & $\checkmark$     & 61.4 & 68.5 & 51.4 & 15.0 & 49.3 & 19.1 \\
%  \bottomrule
% \end{tabular}}
% \vskip -0.05in
% \caption{\textbf{Fine-tuning challenges for image understanding and grounding tasks.} Fine-tuning on one task leads to catastrophic forgetting of the other, while joint fine-tuning remains suboptimal. TWIST \& SCOUT preserves both abilities effectively.}
% \label{tab:finetune_combined}
% \vskip -0.2in
% \end{table*}
Beyond these components, we analyze the impact of $\alpha$-gating, which facilitates knowledge transfer from image understanding to grounding. Instead of learning a standalone grounding module, $\alpha$ controls how much pre-trained features are reused, ensuring the grounding module learns delta features rather than redundant representations. Replacing the learned $\alpha {=} 0.31$ with $\alpha {=} 0$ disrupts this transfer, dropping model performance from 15 mAP to 0, confirming its necessity. These results validate our approach: TWIST ensures knowledge retention, SCOUT enhances grounding, stepwise tuning refines learning efficiency, and $\alpha$-gating enables effective feature reuse across tasks.

% \noindent\textbf{Fine-tuning challenges.} To highlight the limitations of traditional fine-tuning for multimodal tasks, we conduct an ablation comparing simple fine-tuning strategies to our approach in Table~\ref{tab:finetune_combined}. The challenge arises when adapting a pre-trained MLLM for both image understanding and grounding. Image understanding is trained on LLaVA-mix-665k (covering GQA, SQA, and VQA), while grounding uses SCOUT. Fine-tuning solely on SCOUT causes catastrophic forgetting, where the model loses image understanding and generates bounding boxes instead of textual answers, resulting in zero accuracy on GQA, SQA, and VQA. Conversely, fine-tuning only on LLaVA-mix-665k prevents the model from learning grounding, yielding zero AP on COCO. Even training on both datasets simultaneously remains suboptimal, as the model struggles to balance both skills. These results confirm that simple fine-tuning fails to integrate new abilities while preserving existing ones. TWIST \& SCOUT overcomes this by maintaining image understanding while successfully incorporating grounding.  

\noindent\textbf{Fine-tuning challenges.} We analyze the limitations of standard fine-tuning strategies in Table~\ref{tab:finetune_combined}. Adapting a pre-trained MLLM for both image understanding and grounding is challenging—training on LLaVA-mix-665k (GQA, SQA, VQA) preserves image understanding but prevents grounding, while training on SCOUT erases image understanding, causing the model to generate bounding boxes instead of textual answers. Even training on both datasets together remains suboptimal, as the model struggles to balance both tasks. This issue worsens when adding a dataset with a domain shift. To test this, we train MoE-LLaVA in a multi-task setting using both VQA-RAD \cite{li2023llava}, a biomedical VQA dataset, and LLaVA-mix-665k simultaneously. As shown in Table~\ref{tab:biomedical_ft}, MoE-LLaVA suffers a drop across all tasks, failing to generalize between biomedical reasoning and standard visual question answering. In contrast, TWIST \& SCOUT fine-tunes each module separately while preserving pre-trained knowledge, maintaining strong performance across all tasks. These results show that standard fine-tuning struggles to integrate new abilities without degrading existing ones, especially with domain shifts. TWIST \& SCOUT overcomes this by retaining image understanding while incorporating domain-specific reasoning, demonstrating the benefits of a modular, task-adaptive fine-tuning strategy.

\noindent\textbf{Effect of number of experts.} Table~\ref{tab:num_experts} examines the impact of varying experts in the grounding MoE module, evaluated on RefCOCO and RefCOCO+ test-A and test-B splits. A single expert (equivalent to a simple MLP) underperforms compared to multiple experts, reinforcing our choice of MoEs for flexible parameter allocation to meet grounding tasks' computational demands. Increasing experts from 1 to 2 yields substantial gains, validating the need for multiple experts. However, increasing from 2 to 4 provides only marginal improvements while doubling trainable parameters, making it inefficient. Thus, we adopt the 2-expert configuration for the best balance of performance and efficiency.

% In Table \ref{tab:num_experts} we ablate the impact of varying the number of experts in the grounding MoE module, experimenting with configurations that use 2 and 4 experts, corresponding to approximately 1.67B and 3.3B trainable parameters. While increasing the number of experts to 4 results in a slight performance improvement, the gains are not worth the increased computational cost. This diminishing return suggests that beyond a certain point, additional experts do not enhance model performance for the grounding tasks. As a result, we adopt the 2-expert configuration in all our experiments, finding a good balance between performance and efficiency, as reflected in the results across different datasets.

\begin{table}[h]
\centering
\resizebox{1.0\linewidth}{!}{%
\begin{tabular}{cccccc}
\toprule
 & & \multicolumn{2}{c}{\textbf{RefCOCO}} & \multicolumn{2}{c}{\textbf{RefCOCO+}} \\
\cmidrule(r){3-4} \cmidrule(r){5-6} 
Experts & Parameters & test-A & test-B & test-A & test-B \\ \midrule
1 & 0.8B & 79.8 & 71.6 & 78.4 & 60.2 \\
\rowcolor{high}
2 & 1.6B & 90.2 & 80.3 & 87.7 & 71.9 \\
4 & 3.3B & 90.3 & 80.5 & 88.0 & 72.1 \\
\bottomrule
\end{tabular}
}
\vskip -0.05in
\caption{\textbf{Effect of number of experts.} A single expert (MLP) underperforms, validating the need for MoEs. Two experts match four in performance while using half the parameters.}
\vskip -0.1in
\label{tab:num_experts}
\end{table}

\noindent\textbf{Backbone ablation.} We assess TWIST’s flexibility by testing different backbones, as shown in Table~\ref{tab:backbone_ablation}. While TWIST is built on MoE-LLaVA, replacing it with LLaVA still enables effective grounding, achieving 85.7 test-A / 77.2 test-B on RefCOCO and 81.9 test-A / 63.8 test-B on RefCOCO+. Though MoE-LLaVA performs better due to its expert-based design, these results confirm that TWIST is a general framework adaptable to different base models without being architecture-specific.

\begin{table}[h]
\centering
\resizebox{0.9\linewidth}{!}{%
\begin{tabular}{lcccc}
\toprule
 &  \multicolumn{2}{c}{\textbf{RefCOCO}} & \multicolumn{2}{c}{\textbf{RefCOCO+}} \\
\cmidrule(r){2-3} \cmidrule(r){4-5} 
Methods    & test-A & test-B & test-A & test-B \\ \midrule
LLaVA      & 85.7 & 77.2 & 81.9 & 63.8 \\
\rowcolor{high}
MoE-LLaVA  & 90.2 & 80.3 & 87.7 & 71.9 \\
\bottomrule
\end{tabular}
}
\vskip -0.05in
\caption{\textbf{Backbone ablation.} TWIST generalizes across backbones, enabling grounding when replacing MoE-LLaVA with LLaVA.}
\label{tab:backbone_ablation}
\end{table}

\noindent\textbf{Impact of fine-tuning datasets.} Table~\ref{tab:different_dataset} shows the effect of different fine-tuning datasets on TWIST’s performance. Using only Visual Genome (VG) degrades performance (AP: 9.2, $\text{AP}_{50}$: 38.5) due to its noisy annotations. Adding SCOUT improves results (AP: 12.7, $\text{AP}_{50}$: 44.1), while 
% \begin{wraptable}{r}{0.6\linewidth}
% % \vskip -0.1in
% \centering
% \resizebox{1\linewidth}{!}{%
% \begin{tabular}{ccrr}
% \toprule
%  \multicolumn{2}{c}{\textbf{Dataset Type}} & \multicolumn{2}{c}{\textbf{COCO}} \\
%   \cmidrule(r){1-2} \cmidrule(r){3-4}
%     VG & \ourDataset & AP & $\text{AP}_{\text{50}}$ \\ 
%  \midrule
%  $\checkmark$ & \texttimes & 9.2 & 38.5 \\
%  $\checkmark$ & $\checkmark$ & 12.7 & 44.1 \\
%  \rowcolor{high}
%  \texttimes & $\checkmark$ & 15.0 & 49.3 \\ 
%  \bottomrule
% \end{tabular}}
% \caption{\textbf{Impact of fine-tuning datasets.} Adding the visual genome (VG) dataset degrades performance due to noisy labels, while incorporating SCOUT enhances grounding effectiveness.}
% \label{tab:different_dataset}
% \end{wraptable}
training exclusively on SCOUT yields the best performance 
(AP: 15.0, $\text{AP}_{50}$: 49.3). This highlights the importance of high-quality, visually grounded data, with SCOUT providing a cleaner, more informative signal than VG. 
\begin{table}[h]
\centering
\resizebox{0.6\linewidth}{!}{%
\begin{tabular}{ccrr}
\toprule
 \multicolumn{2}{c}{\textbf{Dataset Type}} & \multicolumn{2}{c}{\textbf{COCO}} \\
  \cmidrule(r){1-2} \cmidrule(r){3-4}
    VG & \ourDataset & AP & $\text{AP}_{\text{50}}$ \\ 
 \midrule
 $\checkmark$ & \texttimes & 9.2 & 38.5 \\
 $\checkmark$ & $\checkmark$ & 12.7 & 44.1 \\
 \rowcolor{high}
 \texttimes & $\checkmark$ & 15.0 & 49.3 \\ 
 \bottomrule
\end{tabular}
}
\vskip -0.05in
\caption{\textbf{Impact of fine-tuning datasets.} Adding the visual genome (VG) dataset degrades performance due to noisy labels, while incorporating SCOUT enhances grounding effectiveness.}
\label{tab:different_dataset}
\end{table}

% We ablate to analyze the impact of different fine-tuning datasets on TWIST's performance, since some datasets can be noisy leading to performance degradation of a model. As shown in Table \ref{tab:different_dataset}, using VG dataset degrades performance, with AP dropping to 9.2 and $\text{AP}_{50}$ to 38.5, due to the noise and inconsistencies in VG annotations, but adding SCOUT improves performance, with the best results (AP: 15.0, $\text{AP}_{50}$: 49.3) achieved when training with only SCOUT. This highlights the value of high-quality, visually grounded data provided by SCOUT for effective grounding.

% \begin{table}[h]
% \centering
% % \vskip 0.1in
% % \vskip 0.1in
%   \resizebox{0.6\linewidth}{!}{%
% \begin{tabular}{ccrr}
% \toprule
%  \multicolumn{2}{c}{\textbf{Dataset Type}} & \multicolumn{2}{c}{\textbf{COCO}} \\
%   \cmidrule(r){1-2} \cmidrule(r){3-4}
%     VG & \ourDataset & AP$\uparrow$ & $\text{AP}_{\text{50}}\!\!\uparrow$ \\ 
%  \midrule
%  $\checkmark$ & \texttimes & 9.2 & 38.5 \\
%  $\checkmark$ & $\checkmark$ & 12.7 & 44.1 \\
%  \rowcolor{high}
%  \texttimes & $\checkmark$ & 15.0 & 49.3 \\ 
%  \bottomrule
% \end{tabular}}
% \caption{\textbf{Impact of fine-tuning datasets.} Adding the visual genome dataset degrades performance due to noisy labels, while incorporating SCOUT enhances grounding effectiveness. VG refers to visual genome dataset.}
% \label{tab:different_dataset}
% % \vskip -0.1in
% \end{table}

\noindent\textbf{Scaling properties of SCOUT.} We assess SCOUT’s impact on localization by varying dataset size from 64k to 3M samples, as shown in Table~\ref{tab:scaling_experiment_example}. TWIST’s performance improves steadily, particularly up to 1M samples, after which gains plateau. This saturation occurs because SCOUT inherits CogVLM’s 3-object-per-image limitation, meaning that beyond 512k samples, additional data increases quantity but not diversity in grounding information. Thus, further scaling becomes ineffective, emphasizing dataset quality over sheer volume for improving grounding performance.

% We explore the scalability of SCOUT in Figure \ref{fig:teaser} (b). Leveraging our synthetic data generation pipeline, we experiment with varying dataset sizes ranging from $64\text{k}$ to $3\text{M}$ samples. As seen from the graph, TWIST's performance improves consistently with increase in SCOUT dataset size, particularly up to $1\text{M}$ data points, after which the improvements begin to plateau. This trend highlights the effectiveness of SCOUT in providing high-quality, diverse training samples, outperforming conventional datasets in scaling the model's capabilities for complex grounding tasks.

\begin{table}[h]
\label{tab:scaling}
\centering
\setlength{\tabcolsep}{6pt}
\renewcommand{\arraystretch}{0.9}
\begin{tabular}{lcccccc}
\toprule
Datasets & $64k$ & $128k$ & $256k$ & $512k$ & $1M$ & $3M$  \\
\midrule
PVOC & 0.21 & 0.34 & 0.59 & 0.68 & 0.69 & 0.69 \\
LVIS & 0.20 & 0.38 & 0.61 & 0.65 & 0.67 & 0.67 \\
COCO & 0.10 & 0.12 & 0.13 & 0.14 & 0.15 & 0.15 \\
\bottomrule
\end{tabular}
\vskip -0.05in
\caption{\textbf{Scaling properties of SCOUT} on localization tasks, showing improvements until saturation at 1M samples.}
\label{tab:scaling_experiment_example}
\vskip -0.2in
\end{table}

%% file: sec/conclusion.tex
\section{Conclusion}
\label{sec:conclusion}

% \cs{The conclusion can benefit from a rewrite. There is too much mentioning of step/step-wise, it also emphasizes the contributions again, which is not so interesting.}

% We present TWIST, a Twin-expert \cs{Stepwise} Tuning framework that fine-tunes pre-trained MLLMs without forgetting using a \cs{step-by-step} training signal. To facilitate this learning framework, we introduce SCOUT, a high-quality synthetic dataset with \cs{stepwise} grounded annotations, offering a rich training signal for complex grounding and reasoning tasks. Finally, we develop an evaluation protocol for grounded image captioning capabilities of MLLMs. Together, TWIST \& SCOUT presents a system to upgrade a pre-trained image understanding model to add new visual grounding abilities by only fine-tuning.  \cs{without forgetting?} 

We propose TWIST, a fine-tuning framework that equips pre-trained MLLMs with visual grounding while preserving their image understanding capabilities. By leveraging SCOUT, a high-quality synthetic dataset, our approach enables effective grounding without full model retraining. Through rigorous evaluation, we demonstrate TWIST \& SCOUT’s ability to enhance multimodal reasoning and localization, providing a scalable solution for integrating new skills into MLLMs.

%% file: sec/appendix.tex
% \clearpage
% \setcounter{page}{1}
% \maketitlesupplementary
\appendix
\counterwithin{figure}{section}
\counterwithin{table}{section}
\counterwithin{equation}{section}

% % \section{Appendix}
% \refstepcounter{letteredsection}
% \section*{\theletteredsection\quad Appendix}
% \addcontentsline{toc}{section}{\theletteredsection\quad Appendix}
\section{Supplementary Material}
The supplementary material consists of the following sections: \ref{subsec:dataset_vis} Dataset Visualization, \ref{subsec:samples} Sample Outputs from TWIST \& SCOUT, \ref{subsec:num_params} Number of Parameters and Datasets and \ref{subsec:incontext} In-context Prompts for Mixtral.

% \refstepcounter{letteredsubsection}
% \subsection*{\theletteredsubsection\quad Dataset Visualization and Statistics}
% \addcontentsline{toc}{subsection}{\theletteredsubsection\quad Dataset Visualization and Statistics}
\subsection{Dataset Visualization and Statistics}
\label{subsec:dataset_vis}
% \subsection{Dataset Visualization}

\begin{figure*}
    \centering
\includegraphics[width=1\textwidth]{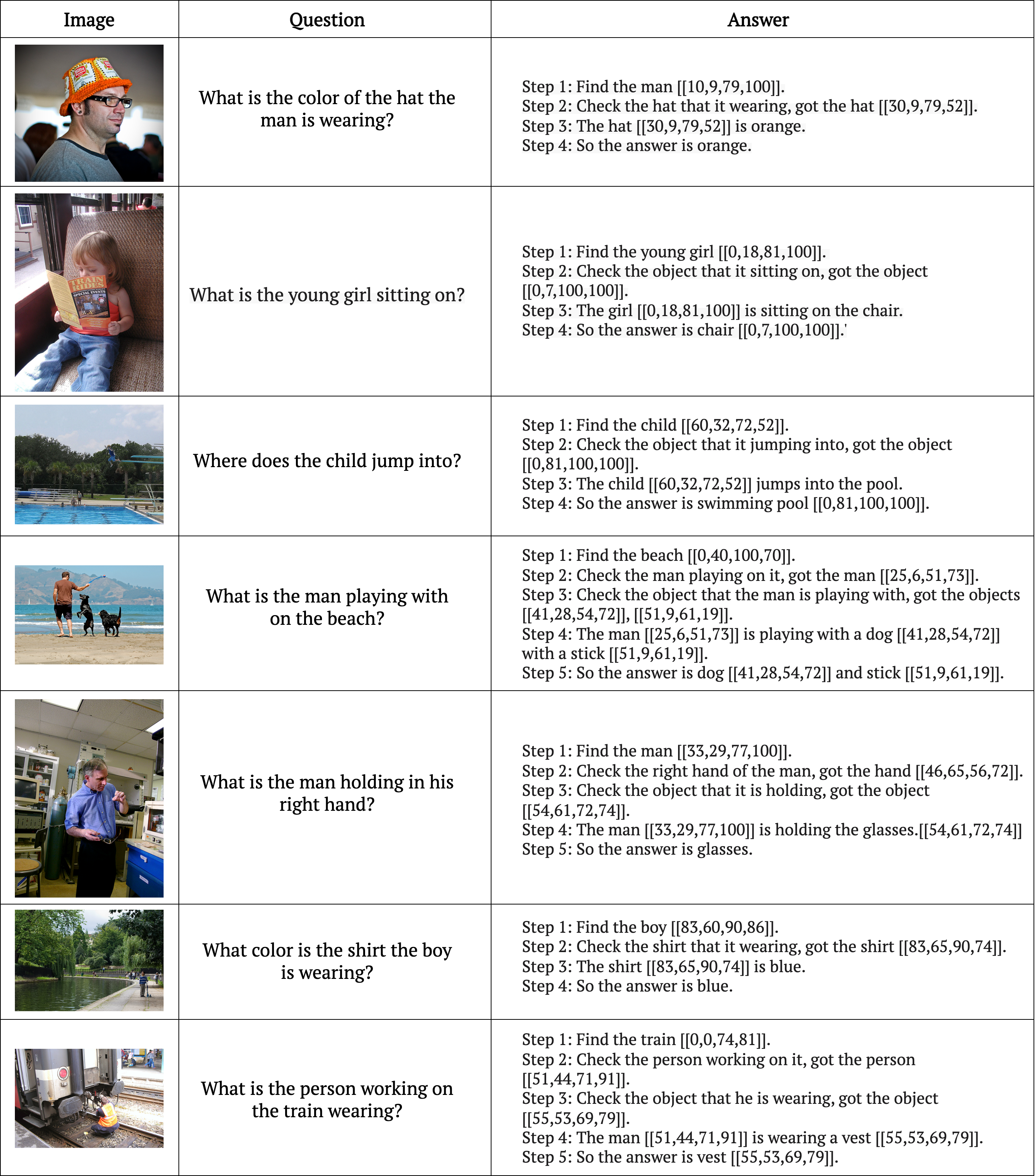}
    \vspace{-3mm}
    \caption{\textbf{Visualization of the Grounded Chain-of-Thought Dataset.} Here we provide step-by-step answers to questions, simplifying the learning process. For instance, identifying the man, finding his hat, determining its color, and finally answering that the hat is orange. This structured approach aids in faster and more effective learning for smaller models.}
    \label{fig:positive}
\end{figure*}

\begin{figure*}
    \centering
\includegraphics[width=1\textwidth]{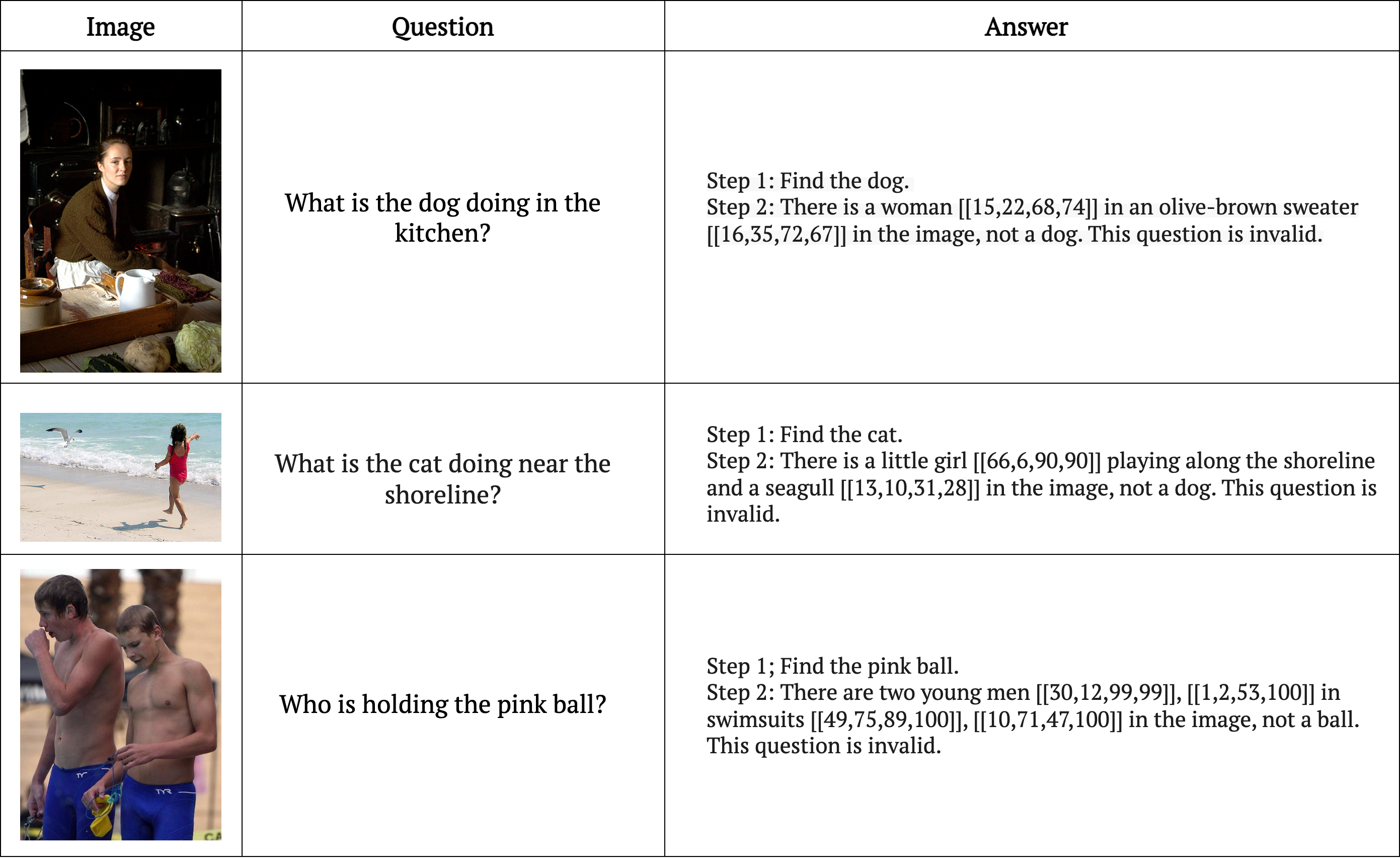}
    \vspace{-3mm}
    \caption{\textbf{Visualization of Negative Samples.} Here we include examples where the question is incorrect or irrelevant, such as asking ``What is the cat doing near the shoreline?" when no cat is present. The model begins by identifying the main object and, if it cannot find the object, declares the question invalid. This negative supervision enhances the model's ability to handle invalid or contradictory queries, improving robustness and accuracy.}
    \label{fig:negative}
\end{figure*}

\begin{figure*}
    \centering
\includegraphics[width=1\textwidth]{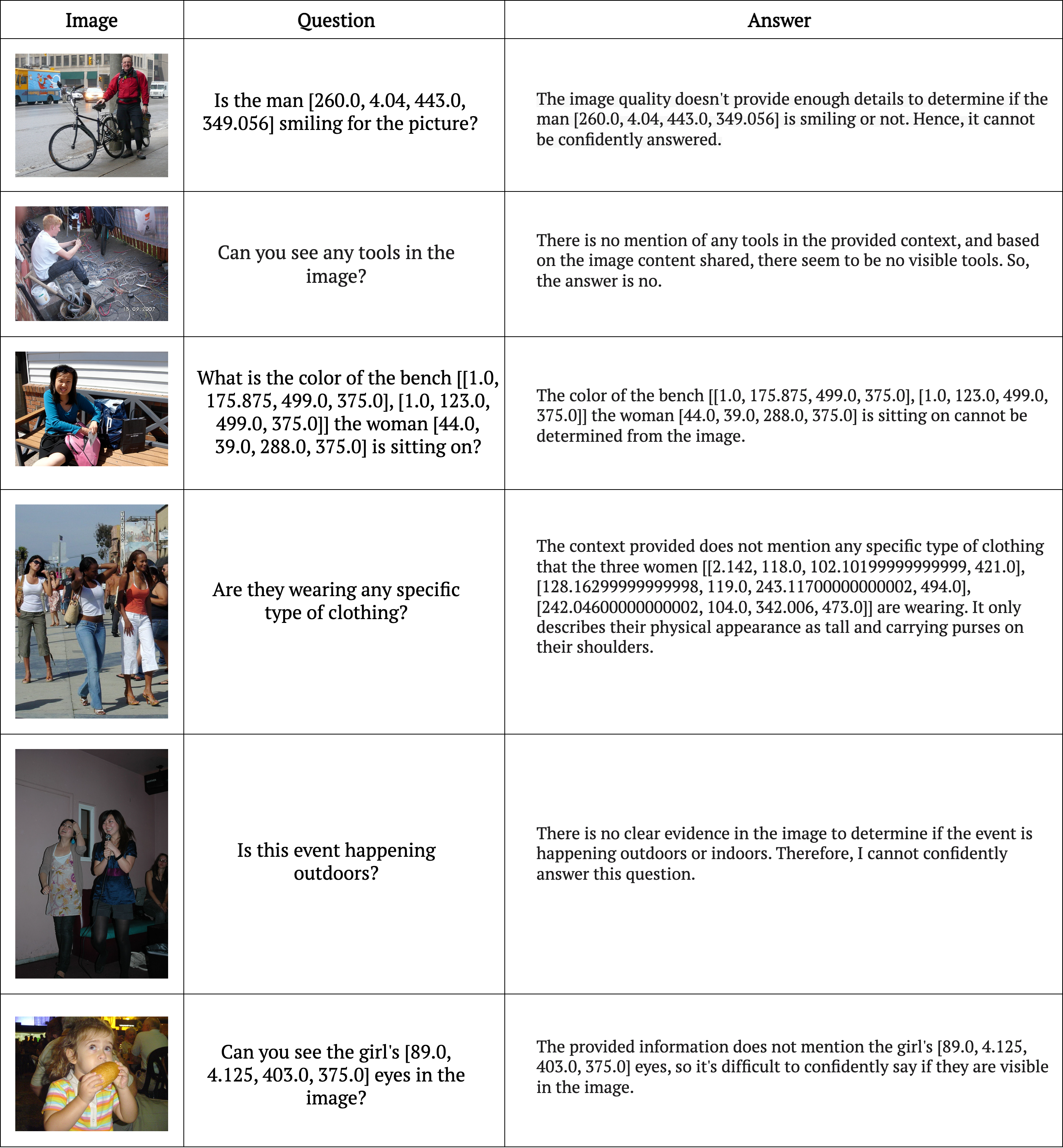}
    \vspace{-3mm}
    \caption{\textbf{Errors in the Grounded Chain-of-Thought data generated by Shikra} due to absence of visual context. In the first row, the LLM fails to determine if the man is smiling, and in the last row, it cannot confirm the visibility of the girl's eyes, despite both being clearly visible in the images. These errors highlight the limitations of relying solely on textual captions for multimodal data generation.}
    \label{fig:shikra}
\end{figure*}

We present dataset statistics comparison between our generated and shikra generated grounded chain-of-though datasets in Table~\ref{tab:dataset_stat} along with three visualizations in this subsection: the positive samples of our synthetic grounded chain-of-thought dataset in Figure \ref{fig:positive}, its negative samples in Figure \ref{fig:negative}, and samples from the noisy dataset generated by Shikra using LLMs in Figure \ref{fig:shikra}.

\noindent\textbf{Dataset Statistics.}
We present a comparison of dataset statistics between our Synthetic Grounded Chain-of-Thought (\ourDataset) dataset and the Shikra-generated dataset in Table \ref{tab:dataset_stat}. The table highlights key metrics such as the number of images, words, turns, objects, and Q/A pairs. This comparison demonstrates the scale and richness of our \ourDataset dataset.

\begin{table}[h]
\centering
\caption{\textbf{Comparison of dataset statistics} between our synthetic data and the Shikra-generated dataset.}
\vskip 0.05in
  \resizebox{\linewidth}{!}{%
% \begin{tabular}{p{1.5cm}|l|cp{0.9cm}p{0.9cm}p{0.9cm}p{0.9cm}p{0.9cm}p{0.9cm}p{0.9cm}c}
\begin{tabular}{lccccc}
\toprule
             & \textbf{Images} & \textbf{Words} & \textbf{Turns} & \textbf{Objects} & \textbf{Q/A Pairs}  \\
\midrule
 Shikra      & $883$     & $7106$    & $1$          & $23692$  & $5922$     \\
 \ourDataset & $30000$   & $15524$   & $\sim4$  & $654314$ & $3113763$     \\
 \bottomrule
 % \vspace*{-1cm}
\end{tabular}}
\label{tab:dataset_stat}
% \vskip -0.1in
\end{table}

\noindent\textbf{\ourDataset: Synthetic Grounded Chain-of-Thought Dataset.}
This dataset is designed to provide step-by-step answers to questions, thereby simplifying the learning process for our models. For example, in the first row of Figure \ref{fig:positive}, when asked "What is the color of the hat the man is wearing?", instead of directly trying to answer the question,  the dataset breaks down the task into manageable steps:
\begin{enumerate}
    \item Identify the man in the image.
    \item Find the hat he is wearing.
    \item Determine the color of the hat.
    \item Provide the final answer: ``The hat is orange."
\end{enumerate}
This structured breakdown helps our smaller models learn more effectively and quickly by reducing the complexity of the task.

\noindent\textbf{Negative Samples.}
In Figure \ref{fig:negative}, we include some negative samples for our \ourDataset dataset, where the question is intentionally incorrect or irrelevant to the image. This is done to mitigate the hallucinations of TWIST \& SCOUT. For instance, in the second row of Figure \ref{fig:negative}, the image shows a girl at the shoreline, but the question asks, ``What is the cat doing near the shoreline?" Our methodology begins by attempting to identify the main object (in this case, the cat). If the model cannot find a cat in the image, it correctly identifies the question as invalid. This type of negative supervision is crucial for training our model to recognize and handle invalid or contradictory queries, thereby improving its robustness and accuracy. 

We generate $40k$ negative samples for our dataset along with the $3M$ positive samples. The whole dataset will be released.
 
\noindent\textbf{Grounded chain-of-thought dataset by Shikra.}
Finally, we visualize samples from the dataset generated by Shikra using LLMs in Figure \ref{fig:shikra}. These examples highlight common errors due to the absence of visual context during data generation. For example, in the first row, the question asks ``Is the man [260.0, 4.04, 443.0, 349.056] smiling for the picture?" and the ground truth response for this in their dataset is ``The image quality doesn't provide enough details to determine if the man [260.0, 4.04, 443.0, 349.056] is smiling or not. Hence, it cannot be confidently answered." We can clearly see from the image that the man is smiling. However, in the Flickr30K dataset, where this image and its captions originate, the captions do not mention that the man is smiling. Shikra uses an LLM to generate data based solely on these captions, without analyzing the image itself. As a result, the LLM states that it cannot determine if the man is smiling because the captions do not provide this information. Similarly, in the last row, the question asks, ``Can you see the girl's [89.0, 4.125, 403.0, 375.0] eyes in the image?" with the ground truth answer stating, ``The provided information does not mention the girl's [89.0, 4.125, 403.0, 375.0] eyes, so it's difficult to confidently say if they are visible in the image." However, the girl's eyes are clearly visible in the image. Such errors demonstrate the limitations of generating datasets without access to visual information and underscore the importance of creating a high-quality dataset like our \ourDataset.

\noindent\textbf{Effect of fine-tuning with negative supervision data.}
The negative supervision samples (Figure \ref{fig:negative}) in \ourDataset offer two key advantages: First, as demonstrated in Table \ref{tab:negative_samples}, training with negative samples leads to improvements in grounded image captioning performance, enhancing the model's ability to distinguish relevant objects.
Second, negative supervision reduces hallucinations, a common issue in MLLMs. For example, as illustrated in Figure \ref{fig:demo}, when faced with a query like ``What is the dog doing near the shoreline?" TWIST \& SCOUT, trained with negative samples, first verifies the presence of the dog before attempting to answer. If no dog is present, and only a girl is in the image, our approach recognizes the question as invalid, thereby avoiding incorrect assumptions and improving overall accuracy.

\begin{table}[h]
\centering
\resizebox{0.75\linewidth}{!}{%
\begin{tabular}{cccc}
\toprule
Positive & Negative & {$\text{AP}\!\uparrow$} & {$\text{AP}_{\text{50}}\!\!\uparrow$} \\
\midrule
$\checkmark$ & \texttimes & 13.8 & 48.1 \\
\rowcolor{high}
$\checkmark$ & $\checkmark$ & 15.0 & 49.3 \\
\bottomrule
\end{tabular}
}
\caption{\textbf{TWIST \& SCOUT with negative samples} show enhanced object detection performance.}
\label{tab:negative_samples}
\end{table}

\subsection{Additional Grounding Comparisons}
\label{sec:appendix_grounding}

To further contextualize the grounding performance of TWIST \& SCOUT, we extend our evaluation by including Grounding DINO \cite{liu2023grounding}  and Ferret \cite{you2023ferret} in both referential expression comprehension (REC) and grounded image captioning tasks. While REC focuses on identifying a single object given a descriptive query, grounded image captioning requires detecting and localizing multiple objects in free-form descriptions, making it a more complex task that integrates both spatial reasoning and image understanding.

\noindent\textbf{Results.} Table~\ref{tab:protocol_extra} shows the grounded image captioning (GIC) results, where TWIST \& SCOUT outperforms all baselines, including PIN and Shikra, and even surpasses Ferret-7B by 2.2 \(\text{AP}_{50}\). Notably, Grounding DINO, a strong object detector, fails entirely at this task, as it lacks the language capabilities required for free-form captioning. While Ferret achieves competitive results, it still falls short of TWIST \& SCOUT, demonstrating the importance of building on pre-trained vision-language models rather than relying solely on extensive pre-training.

On REC tasks in Table~\ref{tab:visual_grounding_extra}, however, Grounding DINO  and Ferret  outperform TWIST \& SCOUT. This is expected, as Grounding DINO is specifically optimized for object detection, and Ferret benefits from large-scale pre-training. However, these models struggle with the more complex grounded captioning task, which requires both visual grounding and high-level reasoning. TWIST \& SCOUT, despite being a fine-tuning approach, effectively balances these two aspects, reinforcing its role as a method for augmenting pre-trained MLLMs with grounding abilities without sacrificing image understanding.

These results highlight a key distinction between our approach and object detection-based models: while object detectors excel at tasks like REC, they cannot generalize to complex grounding tasks that require a fusion of spatial and semantic knowledge. TWIST \& SCOUT, by contrast, enables MLLMs to handle such challenges efficiently without full retraining.

\begin{table}[h]
\centering
\resizebox{\linewidth}{!}{%
\setlength{\tabcolsep}{4pt}
\begin{tabular}{lrrrr}
\toprule
 Method & Parameters & {$\text{AP}\!\uparrow$} &{$\text{AP}_{\text{50}}\!\!\uparrow$} & {$\text{AP}_{\text{L}}\!\uparrow$} \\
 \midrule
 PIN           & 1.2M & 0    & 0    & 0 \\
 Grounding DINO & 172M & 0    & 0    & 0 \\
 Shikra        & 7.0B & 13.2 & 46.8 & 16.7 \\
 Ferret        & 7.0B & 13.9 & 47.1 & 17.4 \\
 \rowcolor{high}
 TWIST \& SCOUT & 1.6B & 15.0 & 49.3 & 19.1 \\
 \midrule
 \rowcolor{palegray}
 Upper bound: CogVLM & 17.0B & 16.1 & 52.7 & 21.3 \\
 \bottomrule
\end{tabular}%
}
\caption{\textbf{Grounded image captioning comparison.} TWIST \& SCOUT outperforms PIN~\cite{dorkenwald2024pin} and Shikra~\cite{chen2023shikra}, demonstrating the benefit of fine-tuning pre-trained MLLMs for complex grounding tasks. While Ferret-7B performs competitively, TWIST \& SCOUT surpasses it by 2.2 \(\text{AP}_{50}\), reinforcing the importance of leveraging existing vision-language knowledge. Grounding DINO, despite excelling in object detection, fails entirely at this task due to its lack of language capabilities. While all models detect fewer objects per image than COCO’s annotations, limiting overall mAP, TWIST \& SCOUT narrows the gap to CogVLM~\cite{wang2023cogvlm}—our upper bound—within $1\%$, highlighting its effectiveness in multi-object grounding.}

\label{tab:protocol_extra}
\vspace{-0.1in}
\end{table}

\begin{table*}[h]
\centering
% \vskip 0.1in
\resizebox{1.0\textwidth}{!}{%
\begin{tabular}{lccccccccccc}
\toprule
 & & \multicolumn{3}{c}{\textbf{RefCOCO}} & \multicolumn{3}{c}{\textbf{RefCOCO+}} & \multicolumn{2}{c}{\textbf{RefCOCOg}} & \multicolumn{2}{c}{\textbf{Flickr30k Ent.}} \\
\cmidrule(r){3-5} \cmidrule(r){6-8} \cmidrule(r){9-10} \cmidrule(r){11-12}
 Method & Type & val & test-A & test-B & val & test-A & test-B & val & test & val & test \\ \midrule
 OFA-L~\citep{wang2022ofa} & pre-trained & 80.0 & 83.7 & 76.4 & 68.3 & 76.0 & 61.8 & 67.6 & 67.6 & -- & -- \\
 VisionLLM-H~\citep{wang2024visionllm} & pre-trained & -- & 86.7 & -- & -- & -- & -- & -- & -- & -- & -- \\
 Shikra-7B~\citep{chen2023shikra} & pre-trained & 87.0 & 90.6 & 80.2 & 81.6 & 87.4 & 72.1 & 82.3 & 82.2 & 75.8 & 76.5 \\
 Grounding DINO~\citep{liu2023grounding} & pre-trained & 90.6 & 93.2 & 88.2 & 82.7 & 88.9 & 75.9 & 86.1 & 87.0 & -- & -- \\
 Ferret-7B~\citep{you2023ferret} & pre-trained & 87.5 & 91.3 & 82.4 & 80.8 & 87.4 & 73.1 & 83.9 & 84.8 & 80.4 & 82.2 \\
 \midrule
 PIN~\citep{dorkenwald2024pin} & fine-tuned & -- & 26.4 & -- & -- & -- & -- & -- & -- & -- & -- \\
 % TWIST & 83.9 & 89.1 & 77.6 & 77.0 & 84.3 & 68.1 & 79.3 & 78.3 & - & - \\
 \rowcolor{high}
 TWIST \& \ourDataset & fine-tuned & 87.2 & 90.2 & 80.3 & 81.6 & 87.7 & 71.9 & 82.6 & 83.1 & 76.8 & 77.9 \\
 \bottomrule
\end{tabular}}
\vskip -0.05in
\caption{\textbf{Visual grounding comparison} on the REC task. Grounding DINO and Ferret outperform TWIST \& SCOUT in referential expression comprehension (REC), as expected, since they are optimized for object detection and benefit from large-scale pre-training. However, these models struggle in grounded image captioning, where TWIST \& SCOUT excels by leveraging both spatial reasoning and image understanding. PIN, as a fine-tuning approach, falls behind across all tasks. Note that ``--'' means the numbers are not reported by the baselines.}

\label{tab:visual_grounding_extra}
\vskip -0.1in
\end{table*}

\subsection{Sample Outputs from TWIST \& SCOUT}
\begin{figure*}
    \centering
\includegraphics[width=0.85\textwidth]{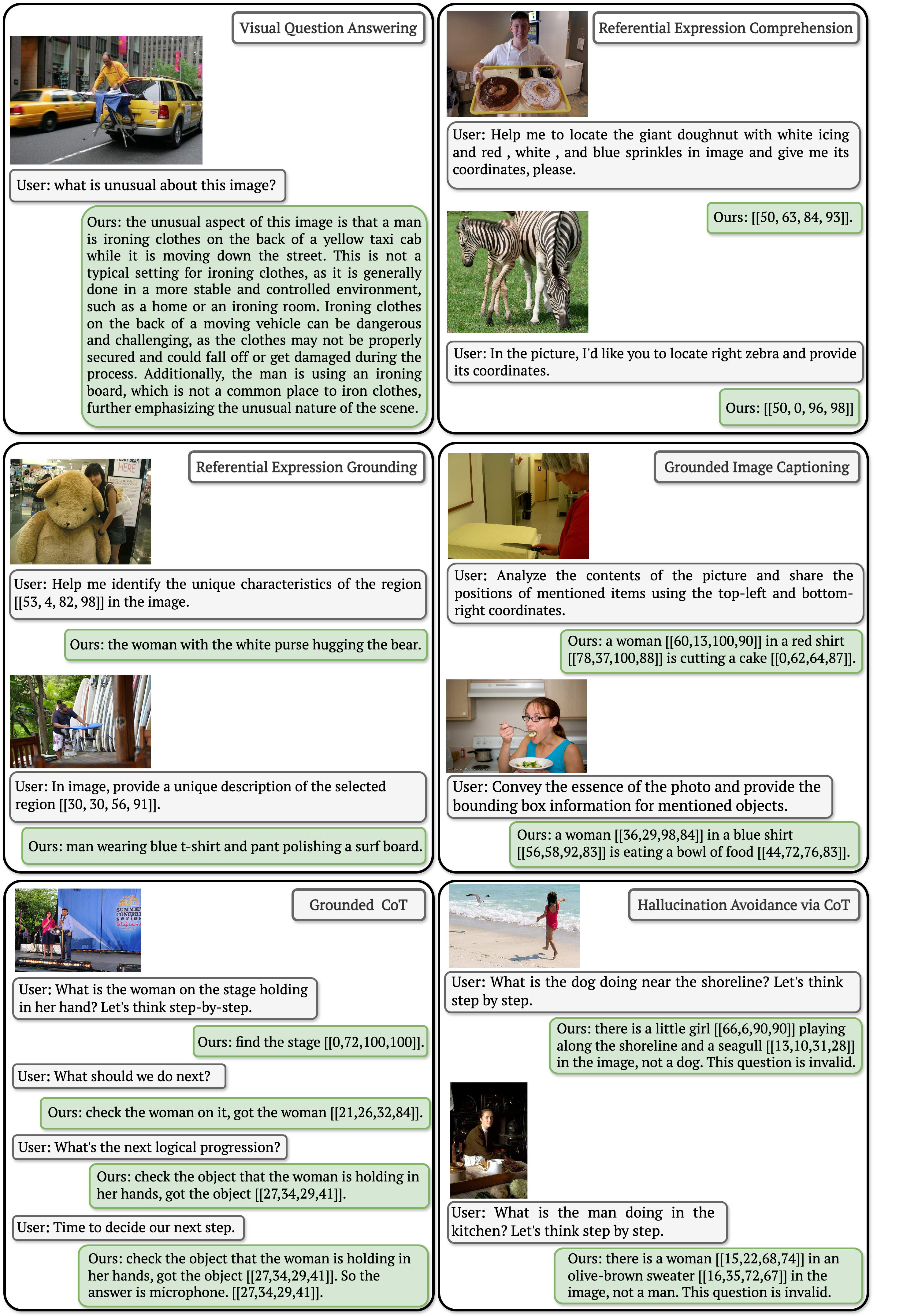}
    \vspace{-3mm}
    \caption{\textbf{Samples generated by TWIST \& SCOUT trained on our \ourDataset.}}
    \label{fig:demo}
\end{figure*}

% \refstepcounter{letteredsubsection}
% \subsection*{\theletteredsubsection\quad Sample Outputs from LynX}
% \addcontentsline{toc}{subsection}{\theletteredsubsection\quad Sample Outputs from LynX}
\label{subsec:samples}
In Figure \ref{fig:demo}, we showcase outputs generated by our TWIST \& SCOUT model trained on \ourDataset. The image demonstrates our model's versatility across a wide range of tasks, including visual question answering, referential expression comprehension, referential expression grounding, grounded image captioning, and grounded chain of thought. Additionally, TWIST \& SCOUT effectively avoids hallucination through its chain-of-thought reasoning.

\subsection{Number of Parameters and Datasets}
\label{subsec:num_params}
In table \ref{table:params_datasets_numbers}, we provide the number of trainable parameters of each baseline and the training dataset used for each baseline model used for our paper. As seen, compared the our baselines, TWIST \& SCOUT efficiently achieves competitive performance with only 1.67 billion trainable parameters and 0.8 billion active parameters, which is significantly less than most models with similar capabilities. Although we have more parameters than PIN, it is limited to generating single bounding box locations per prompt and cannot perform other tasks. Moreover, TWIST \& SCOUT accomplishes this feat while utilizing a relatively modest training dataset of 651k image-caption pairs, showcasing its ability to extract maximum value from limited data and potentially offering improved scalability and resource efficiency compared to methods requiring billions of parameters or massive training datasets. 

\begin{table*}[t]
\caption{Comparison of image-caption models: trainable parameters, active parameters, and training dataset sizes. TWIST \& SCOUT achieves competitive performance with fewer parameters and a smaller dataset compared to most models with similar capabilities.}
\centering
\begin{tabular}{lccll}
\toprule
\textbf{Method} & \textbf{Trainable Parameters} & \textbf{Active Parameters} & \textbf{Size of Training Dataset} \\
\midrule
PIN & 1.4M & 1.4M & 70k image-caption pairs \\
Shikra & 7B (13B) & 7B (13B) & 7.8M image-caption pairs \\
CogVLM & 17B & 17B & 1B image-caption pairs \\
OFA-L* & 470M & 470M & 24.37M image-caption pairs \\
VisionLLM-H & 1.62B & 1.62B & 738k image-caption pairs \\
I-80B & 80B & 80B & 300M image-caption pairs \\
LLaVA-1.5 & 13B & 13B & 7.8M image-caption pairs \\
\rowcolor{high}
TWIST \& SCOUT & 1.67B & 800M & 651k image-caption pairs \\
\bottomrule
\end{tabular}
\label{table:params_datasets_numbers}
\end{table*}

\subsection{In-context Prompts for Mixtral}
\begin{figure*}
    \centering
\includegraphics[width=1\textwidth]{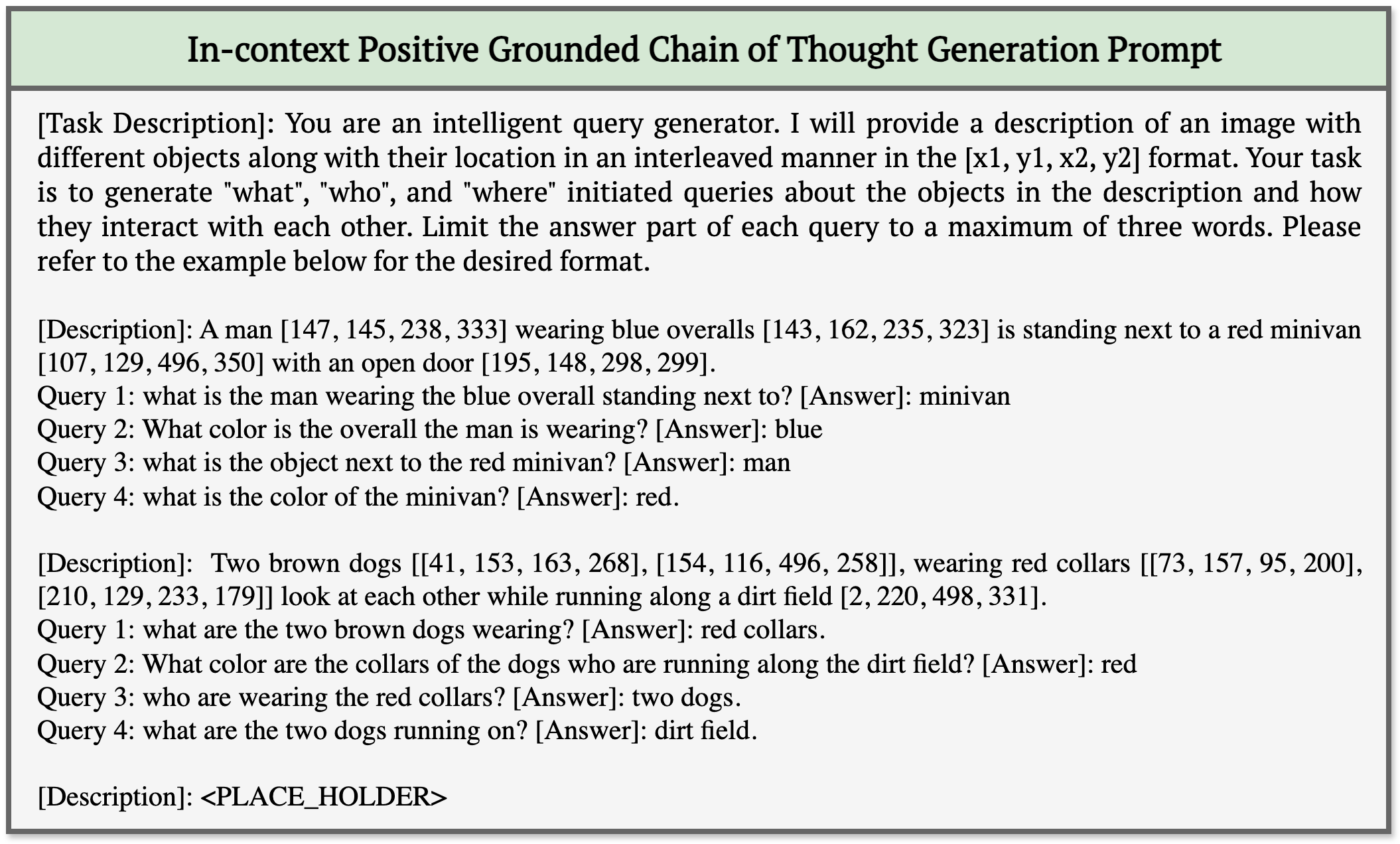}
    \vspace{-3mm}
    \caption{\textbf{Prompt used for generating engaging and relevant questions for the positive samples in our \ourDataset dataset}, demonstrating Mixtral's ability to enhance query formulation.}
    \label{fig:positve_prompts}
\end{figure*}

\begin{figure*}
    \centering
\includegraphics[width=1\textwidth]{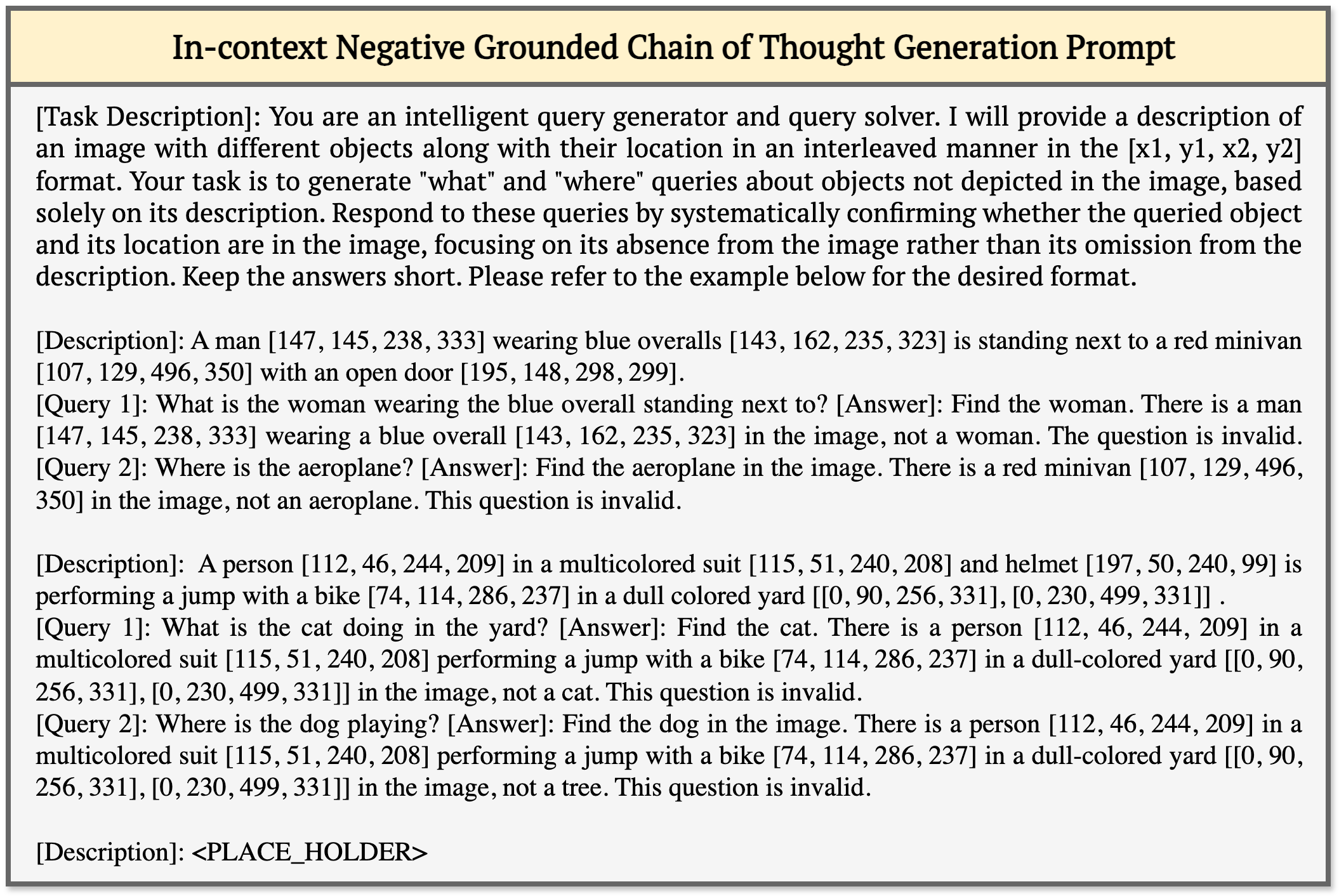}
    \vspace{-3mm}
    \caption{\textbf{Prompt utilized for creating negative samples in our \ourDataset dataset}, showcasing the method for generating queries that highlight contradictions or irrelevant information.}
    \label{fig:negative_prompts}
\end{figure*}

\begin{figure*}
    \centering
\includegraphics[width=1\textwidth]{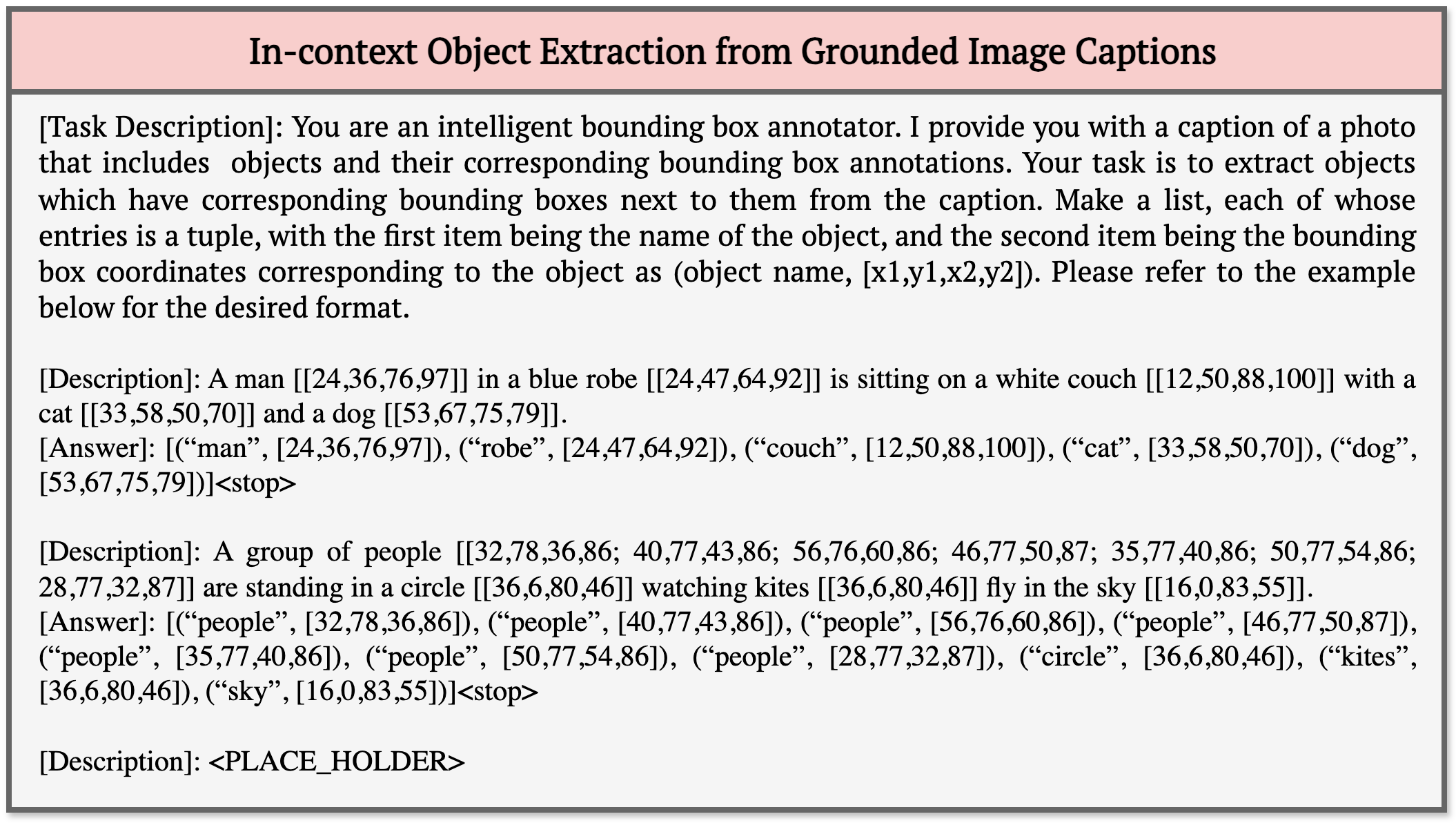}
    \vspace{-3mm}
    \caption{\textbf{Prompt used to extract object names and bounding boxes} from grounded image captions generated by our VLM models, illustrating the process of transforming free-form text outputs into structured data.}
    \label{fig:parse}
\end{figure*}

% \refstepcounter{letteredsubsection}
% \subsection*{\theletteredsubsection\quad In-context Prompts for Mixtral}
% \addcontentsline{toc}{subsection}{\theletteredsubsection\quad In-context Prompts for Mixtral}
\label{subsec:incontext}
Large Language Models (LLMs) excel in in-context learning scenarios, where they can understand and perform tasks based on provided examples within the input context. This ability allows LLMs to adapt to various tasks without requiring explicit retraining. By leveraging patterns and information from the input context, LLMs can generate coherent and relevant responses, making them highly versatile and effective across diverse applications. Leveraging this quality, we employ Mixtral, an open-source LLM, to generate queries, a crucial step in creating our \ourDataset dataset. Additionally, we utilize this LLM's ability to extract object names and bounding boxes from the free-form text outputs of our VLM models, particularly in grounded image captioning. Figure \ref{fig:positve_prompts} illustrates the prompt used for generating interesting questions for the positive samples in our \ourDataset dataset. Figure \ref{fig:negative_prompts} shows the prompts used to generate negative samples. Finally, Figure \ref{fig:parse} depicts the prompts employed to extract objects from grounded image captions produced by our models.

